\documentclass{article}

\usepackage[final,nonatbib]{neurips_2022}
\usepackage[utf8]{inputenc} 
\usepackage[T1]{fontenc}    
\usepackage{hyperref}       
\usepackage{url}            
\usepackage{booktabs}       
\usepackage{amsfonts}       
\usepackage{nicefrac}       
\usepackage{microtype}      
\usepackage{xcolor}         
\usepackage[square,numbers,sort&compress]{natbib} 

\usepackage{graphicx}
\usepackage{subfigure}
\usepackage{booktabs} 
\usepackage{amsmath, amsthm, amsfonts, amssymb}

\usepackage{algorithmic, algorithm}

\newtheorem{theorem}{Theorem}
\newtheorem{assume}{Assumption}

\usepackage{verbatim}
\usepackage{mathtools}
\usepackage{bbm}
\usepackage{authblk}

\title{Beyond accuracy: generalization properties of bio-plausible temporal credit assignment rules}

\author[1,2,3,*]{Yuhan Helena Liu}
\author[4,5]{Arna Ghosh}
\author[4,5,6,7]{Blake A. Richards}
\author[1,2,3]{Eric Shea-Brown}
\author[5,7,8,*]{Guillaume Lajoie}
\affil[1]{Department of Applied Mathematics, University of Washington, Seattle, WA, USA}
\affil[2]{Allen Institute for Brain Science, 615 Westlake Ave N, Seattle WA, USA}
\affil[3]{Computational Neuroscience Center, University of Washington, Seattle, WA, USA}
\affil[4]{School of Computer Science, McGill University, Montreal, QC, Canada}
\affil[5]{Mila - Quebec AI Institute, Montreal, QC, Canada}
\affil[6]{Department of Neurology and Neurosurgery, Montreal Neurological Institute, McGill University, Montreal, QC, Canada}
\affil[7]{Canada CIFAR AI Chair, CIFAR, Toronto, ON, Canada}
\affil[8]{Dept. de Mathématiques et Statistiques, Université de Montréal, Montreal, QC, Canada}
\affil[*]{Correspondence: hyliu24@uw.edu, g.lajoie@umontreal.ca}

\begin{document}
	
	\maketitle
	
	\begin{abstract}
		
		To unveil how the brain learns, ongoing work seeks biologically-plausible approximations of gradient descent algorithms for training recurrent neural networks (RNNs). Yet, beyond task accuracy, it is unclear if such learning rules converge to solutions that exhibit different levels of generalization than their non-biologically-plausible counterparts. Leveraging results from deep learning theory based on loss landscape curvature, we ask: how do biologically-plausible gradient approximations affect generalization?  We first demonstrate that state-of-the-art biologically-plausible learning rules for training RNNs exhibit worse and more variable generalization performance compared to their machine learning counterparts that follow the true gradient more closely. Next, we verify that such generalization performance is correlated significantly with loss landscape curvature, and we show that biologically-plausible learning rules tend to approach high-curvature regions in synaptic weight space. Using tools from dynamical systems, we derive theoretical arguments and present a theorem explaining this phenomenon. This predicts our numerical results, and explains why biologically-plausible rules lead to worse and more variable generalization properties. Finally, we suggest potential remedies that could be used by the brain to mitigate this effect. To our knowledge, our analysis is the first to identify the reason for this generalization gap between artificial and biologically-plausible learning rules, which can help guide future investigations into how the brain learns solutions that generalize.
		
	\end{abstract}
	
	\section{Introduction}
	
	A longstanding question in neuroscience is how animals excel at learning complex behavior involving temporal dependencies across multiple timescales and thereafter generalize this learned behavior to unseen data. This requires solving the temporal credit assignment problem: how to assign the contribution of past neural states to future outcomes. To address this, neuroscientists are increasingly turning to the mathematical framework provided by recurrent neural networks (RNNs) to model learning mechanisms in the brain~\cite{lillicrap2019backpropagation,prince2021ccn}. Temporal credit assignment in RNNs is typically achieved by backpropagation through time (BPTT), or other gradient descent-based optimization algorithms, none of which are biologically-plausible (or bio-plausible for short). Therefore, the use of RNNs as a framework to understand the computational principles of learning in the brain has motivated an influx of bio-plausible learning rules that approximate gradient descent~\cite{lillicrap2019backpropagation,Marschall2019,prince2021ccn}.
	
\begin{figure}[h!]
		\centering
		\includegraphics[width=0.99\textwidth]{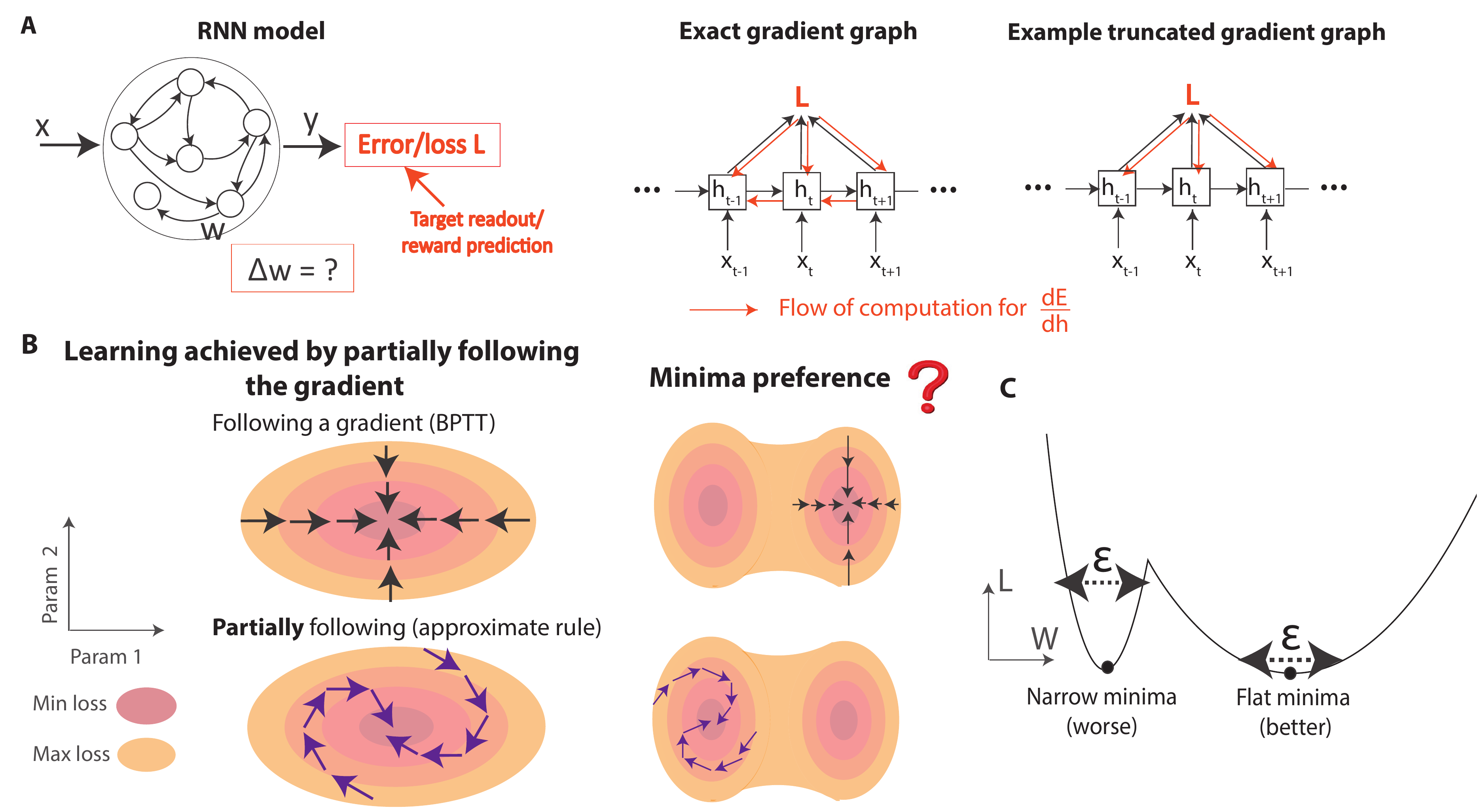}
		\caption{\textbf{Setup}. A) Illustration of an RNN trained to minimize error/loss function L (left). Existing bio-plausible proposals for RNNs estimate the gradient by neglecting dependencies that are biologically implausible to compute (right). B) Low training error/loss can be achieved by \textbf{partially} following a gradient (right), but the preference for converging to minima with certain generalization properties remains underexamined for these learning rules (left). C) Minima flatness matters: 1-D loss landscape illustration with two solutions that equally minimize loss $L$, but exhibit drastically different generalization properties: the narrower minima are more sensitive to perturbation. 
		} 
		\label{fig:Figure1}
	\end{figure}

	The performance of such rules is typically quantified by accuracy. Although the accuracy achieved by these rules is often comparable to true gradient descent, little is known about the breadth of the emergent solutions, namely how well they generalize. Broadly speaking, generalization refers to a trained model's ability to adapt to previously unseen data, and is typically measured by the so-called \textbf{generalization gap}: the difference between training and testing error. This is especially important when learning complex tasks with nonlinear RNNs where the loss landscape is non-convex, and therefore, many solutions with comparable training accuracy can exist. These solutions, characterized as (local) minima in the loss landscape, can nonetheless exhibit drastically different levels of generalization (Figure~\ref{fig:Figure1}). It is not clear if gradient-based methods like BPTT and the existing bio-plausible alternatives have a different tendency to converge to loss minima that provide better or worse generalization. 
	
	While the search for better predictors of the generalization performance remains an open issue in deep learning research~\cite{jiang2020neurips}, recent extensive studies identify flatness of the loss landscape at the solution point as a promising predictor for generalization~\cite{jiang2019fantastic,petzka2021relative,dziugaite2017computing,neyshabur2017exploring,tsuzuku2020normalized}. Leveraging these empirical and theoretical findings, \textbf{we ask}: how do proposed biologically-motivated gradient approximations affect the flatness of the converged solution on loss landscape, and thereby, generalization? 
	
	Our overarching goal is to investigate generalization trends for existing bio-plausible temporal credit assignment rules, and in particular, examine how truncation-based bio-plausible gradient approximations can affect such trends. Specifically, \textbf{our contributions} are summarized as follows:
	\begin{itemize}
		\item In numerical experiments, we demonstrate across several well-known neuroscience and machine learning benchmark tasks that state-of-the-art (SoTA) bio-plausible learning rules for training RNNs exhibit worse and more variable generalization gaps, compared to true (stochastic) gradient descent (Figure~\ref{fig:Figure2}A-C).
		\item Using the same experiments, we show that bio-plausible learning rules tend to approach high-curvature regions in synaptic weight space as measured by the loss’ Hessian eigenspectrum (Figure~\ref{fig:Figure2}D-F). Further, we verify that this correlates with worse generalization (Figure~\ref{fig:Figure2}D-F and~\ref{fig:LamCurves}), which is consistent with the literature.
		\item We present a theorem to explain this phenomenon by examining the weight update equation as a discrete dynamical system, which sheds light on a potential connection between gradient alignment and the preference over converging to narrower minima (Theorem~\ref{thm:rank1_eval}, Figure~\ref{fig:LamCurves_rholr}). 
	\end{itemize}
	Given the core components in designing artificial neural networks: data, objective functions, learning rules and architectures~\cite{richards2019deep}, we investigate different learning rules while holding the data, objective function and architecture constant. SoTA RNN learning rules investigated include a three-factor rule with symmetric feedback (symmetric e-prop~\cite{Bellec2020}), a three-factor rule using random feedback weights (RFLO~\cite{murray2019local}) and a multi-factor rule using local modulatory signaling (MDGL)~\cite{liu2021pnas}. For an in-depth explanation of how these rules are implemented and why they are bio-plausible, please refer to Appendix~\ref{scn:net_details}. We also encourage the reader to visit Appendix B for Theorem~\ref{thm:rank1_eval} proof and discussion on loss landscape geometry. In the last paragraph of the Discussion section, we discuss potential remedies implemented by the brain and provide preliminary results (Appendix Figure~\ref{fig:LamCurves_lrdecay}). To our knowledge, our analysis is the first to highlight and quantitatively provide a mechanistic explanation of the reason for this gap in solution quality between artificial and bio-plausible learning rules for RNNs, thereby motivating further investigations into how the brain learns solutions that generalize.
	
	\section{Related works}
	
	\subsection{Bio-plausible gradient approximations}
	
	Investigating bio-plausible learning rules is of interest both to identify more efficient training strategies for artificial networks and to better understand learning in the brain~\cite{bredenberg2021impression,bahroun2021normative,confavreux2020meta,ocker2021tensor,tyulmankov2021biological,meulemans2021credit,prince2021ccn,Marschall2019,mcmahan2021learning,haider2021latent,pogodin2021towards,friedrich2021neural,bredenberg2020learning,illing2021local,najarro2020meta,tano2020local,duncker2020organizing,nayebi2020identifying,ashwood2020inferring,golkar2020simple,kepple2021curriculum,tyulmankov2022meta,moore2020using,ang2021functional,sezener2021rapid,marschall2019using,clark2021credit,ernoult2022towards,lindsey2020learning,lim2015inferring,luczak_neurons_2022,whittington_theories_2019}. In order for learning to reach a certain goal quantified by an objective, learning algorithms often minimize a loss function~\cite{richards2019deep}. The error gradient, if available, tells us how each parameter should be adjusted in order to have the steepest local descent. For training RNNs, which are widely used as a model for neural circuits~\cite{vyas2020computation,perich2021inferring,smith2021reverse,kleinman2021mechanistic,schuessler2020interplay,schaeffer2020reverse,yang2019task,mante2013context,glaser2020recurrent,kadmon2020predictive,dong2020reservoir,turner2021charting,nestler2020unfolding,kim2021strong,valente2021probing,macpherson2021natural,tsuda2022neuromodulators,rungratsameetaweemana2022probabilistic,depasquale2018full,langdon2022latent}, standard algorithms that follow this gradient --- real time recurrent learning (RTRL) and BPTT --- are not bio-plausible and have overwhelming memory storage demands~\cite{Williams1995,Marschall2019}. However, learning rules that only approximate the true gradient can sometimes be as effective as those that follow the gradient exactly~\cite{richards2019deep,linsley2020}. Because of that, bio-plausible learning rules that approximate the gradient using known properties of real neurons have been proposed and led to successful learning outcomes across many different tasks in feedforward networks~\cite{lillicrap2020backpropagation,lillicrap2019backpropagation,pozzi2018biologically,laborieux2021scaling,millidge2020activation,sacramento2018dendritic,amit2019deep,rubin2021credit,payeur2020burst,Roelfsema2018,lillicrap2016random}, with recent extensions to recurrently connected networks~\cite{murray2019local,Bellec2020,liu2021pnas}. These existing bio-plausible rules for training RNNs~\cite{murray2019local,Bellec2020,liu2021pnas} are truncation-based (which is the focus of this study), so that the untruncated gradient terms can be assigned with putative identities to known biological learning ingredients: eligibility traces, which maintain preceding activity on the molecular levels~\cite{Magee2020,Gerstner2018,Sanhueza2013,Cassenaer2012,Yagishita2014,Suvrathan2019}, combined with top-down instructive signaling~\cite{Gerstner2018,Magee2020,schultz2016dopamine,aljadeff2019cortical,florian2007reinforcement,Farries2007,legenstein2008learning,brzosko2019neuromodulation,pawlak2010modstdp} as well as local cell-to-cell modulatory signaling within the network~\cite{smith2019single,smith2020new,liu2021pnas}. For efficient online learning in RNNs, other approximations (not necessarily bio-plausible) to RTRL~\cite{mujika2018approximating,tallec2017unbiased,cooijmans2019variance,roth2018kernel,Menick2020,zenke2020spike} have also demonstrated to produce a good performance. Given the impressive accuracy achieved by these approximate rules, several studies began to investigate their convergence properties~\cite{cao2020characterizing}, e.g. for random backpropagation weights in feedforward networks~\cite{song2021convergence,girotti2021convergence}. However, the trend of generalization capabilities of these rules, especially in RNNs, is underexamined. 
	
	\subsection{Loss landscape curvature and generalization performance}
	Given the central importance of understanding how neural networks perform in situations unseen during training~\cite{xu2012robustness,jiang2020neurips,allen2019can,advani2020high,jacot2018neural,thomas2020interplay,pezeshki2021gradient,baratin2021implicit}, the deep learning community has made tremendous efforts to develop tools for understanding generalization that we leverage here. 
	That flat minima could lead to better generalization was observed more than two decades ago~\cite{hochreiter1997flat}. Intuitively, under the same amount of perturbation $\epsilon$ in parameter space (e.g. loss landscape changes due to the addition of new data) worse performance degradation will be seen around the narrower minima (see Figure~\ref{fig:Figure1}C). We note that perturbations in parameter space can be linked to that in the input space~\cite{petzka2021relative}. Recently, many empirical studies have consistently supported the usefulness of this predictor~\cite{foret2020sharpness,chaudhari2019entropy,sun2020exploring,keskar2016large,li2018visualizing,feng2021inverse,novak2018sensitivity,wang2018identifying}. In particular, the authors of~\cite{jiang2019fantastic} performed an extensive study and found that flatness-based measures have a higher correlation with generalization than other alternatives. Motivated by this, several studies have characterized properties of the loss functions's Hessian --- whose eigenspectrum carries information about curvature~\cite{sagun2017empirical,ghorbani2019investigation,yao2020pyhessian,liao2021hessian,xie2022power}. Connections between flatness and generalization performance have shed light on the reason for greater generalization gaps in large batch training~\cite{keskar2016large,yao2018hessian,xie2020diffusion,jastrzkebski2018finding,zhang2018theory}, and also have inspired optimization methods to favor flatter minima~\cite{hoffer2017train,foret2020sharpness,chaudhari2019entropy,izmailov2018averaging,baldassi2020shaping,kaddour2022questions,zheng2021regularizing,wu2020adversarial,karakida2019normalization,sankar2020deeper}. Despite the criticism of scale-dependence of flatness~\cite{dinh2017sharp}, where parameter rescaling can drastically change flatness but not always generalization quality, flatness  --- with parameter scales taken into account~\cite{liang2019fisher,rangamani2019scale} --- are connected to PAC-Bayesian generalization error bounds~\cite{dziugaite2017computing,neyshabur2017exploring,tsuzuku2020normalized}. Moreover, a recent theoretical study rigorously connects the flatness of the loss surface to generalization in classification tasks under the assumption that labels are (approximately) locally constant~\cite{petzka2021relative}. Leveraging the great progress from the deep learning community, we aim to study the generalization properties of bio-plausible learning rules from a geometric perspective. 
	
	\section{Results}
	
\begin{figure}[h!]
		\centering
		\includegraphics[width=0.99\textwidth]{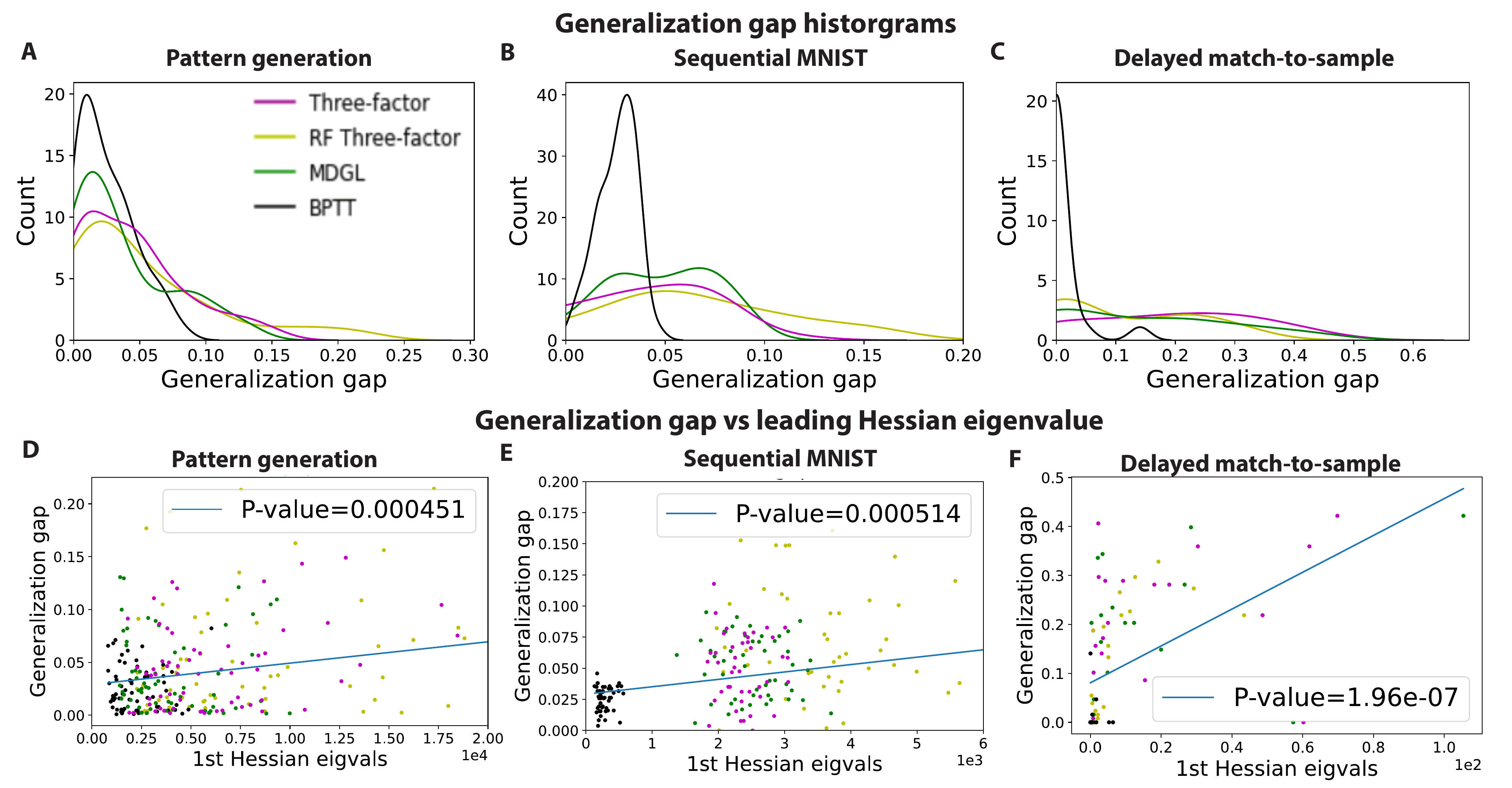}
		\caption{\textbf{Bio-plausible temporal credit assignment rules show worse and more variable generalization gap, which can be informed by loss landscape curvature}. A-C) Generalization gap distributions computed at the end of training across different random weight initializations for several well-known neuroscience and machine learning tasks. The higher the generalization gap, the worse the generalization performance. BPTT (black), bio-plausible alternatives (magenta, yellow and green). D-F) Scatter plots showing the trend of generalization gap v.s. leading loss Hessian eigenvalue across many runs; each point corresponds to a single run of the same runs as in A-C.
		} 
		\label{fig:Figure2}
	\end{figure}

	In this section, we first describe the network and learning setup we use (Figure~\ref{fig:Figure1}A). Next, we present a number of numerical experiments where we compute the generalization gap directly on three commonly used ML and neuroscience tasks (Figure~\ref{fig:Figure2}A-C), for truncation-based bio-plausible gradient approximations. For the same experiments, we also quantify loss landscape curvature along learning trajectories, and connect these quantities to generalization behavior (Figures~\ref{fig:Figure2}D-F and~\ref{fig:LamCurves}). Finally, we provide theoretical arguments and a theorem that explains how gradient alignment in bio-plausible gradient approximations can affect curvature preference (Theorem~\ref{thm:rank1_eval}, Figure~\ref{fig:LamCurves}) and thus, generalization. Through additional experiments, we verify the predictive power of our theory. We conclude with discussions on potential remedies used by the brain (Appendix Figure~\ref{fig:LamCurves_lrdecay}) and future directions.
	
	\subsection{Network and learning setup} \label{scn:setup}
	
	The detailed governing equations of our setup can be found in Methods (Appendix~\ref{scn:methods}). We consider a RNN with $N_{in}$ input units, $N$ hidden units and $N_{out}$ readout units (Figure~\ref{fig:Figure1}A). We verified that trends hold for different network sizes and refer the reader to Appendix~\ref{scn:sim_details} for more details. The update formula for $h_t \in \mathbb{R}^N$ (the hidden state at time $t$) is governed by:
	\begin{equation}
		h_{t+1}= \phi(W_h f(h_t), W_x x_t), \label{eqn:vrnn}
	\end{equation}
	where $\phi(\cdot) : \mathbb{R}^N \rightarrow \mathbb{R}^N$ is the hidden state update function, $f(\cdot) : \mathbb{R}^N \rightarrow \mathbb{R}^N$ is the activation function, $W_h \in \mathbb{R}^{N \times N}$ (resp. $W_x \in \mathbb{R}^{N_{in} \times N}$) is the recurrent (resp. input) weight matrix and $x \in \mathbb{R}^{N_{in}}$ is the input. For $\phi$, we consider a discrete-time implementation of a rate-based recurrent neural network (RNN) similar to the form in~\cite{ehrlich2021psychrnn} (details in Appendix~\ref{scn:methods}). Readout $\hat y \in \mathbb{R}^{N_{out}}$, with readout weights $w \in \mathbb{R}^{N_{out} \times N}$, is defined as
	\begin{equation}
		\hat y = \langle w, f(h_t)\rangle. 
	\end{equation}
	
	We performed experiments on three tasks: sequential MNIST~\cite{lecun1998mnist}, pattern generation~\cite{nicola2017supervised} and delayed match-to-sample tasks~\cite{meyer2011stimulus}.
	The objective is to minimize scalar loss $L \in \mathbb{R}$, which is defined as
	\begin{equation}
		L(W_h) = \begin{cases}
			\frac{1}{2T B}\sum_{i=1}^B \sum_{t=1}^T \sum_{k=1}^{N_{out}} (\hat y^{(i)}_{k,t} - y^{(i)}_{k,t})^2, \text{ for regression tasks} \cr
			\frac{-1}{T B} \sum_{i=1}^B \sum_{t=1}^T \sum_{k=1}^{N_{out}} \pi^{(i)}_{k,t} log \hat \pi^{(i)}_{k,t}, \text{ for classification tasks} 
		\end{cases} \label{eqn:loss_fcn0}
	\end{equation}
	given target readout $y \in \mathbb{R}^{N_{out}}$, task duration $T \in \mathbb{R}$ and batch size $B \in \mathbb{R}$. $\pi_{k,t} \in \mathbb{R}$ is the one-hot encoded target for readout unit $k$ at time $t$ and $\hat \pi_{k,t} = \text{softmax}_k (\hat y_{1,t}, \dots, \hat y_{N_{OUT},t}) = \exp(\hat y_{k,t})/\sum_{k'} \exp(\hat y_{k',t})$ is the predicted category probability.
	
	Different learning algorithms examined in this work are BPTT (our benchmark), which update weights by computing the exact gradient ($\nabla L(W_h) \in \mathbb{R}^{N \times N}$):
	\begin{equation}
		\Delta W_h = -\eta \nabla L(W_h),
	\end{equation}
	and three SoTA bio-plausible learning rules that update weights using approximate gradient:
	\begin{equation}
		\widehat{\Delta W_h} = -\eta \tilde \nabla L(W_h),
	\end{equation}
	where $\tilde \nabla L(W_h) \in \mathbb{R}^{N \times N}$ denotes a gradient approximation and $\eta \in \mathbb{R}$ denotes the learning rate. These three learning rules are explained further in Appendix~\ref{scn:methods} (Methods) but we note that these bio-plausible learning rules are based on truncations of dependency paths --- on the computational graph for the exact gradient--- that are biologically implausible to compute (Figure~\ref{fig:Figure1}A). In all figures, learning rules are labeled as "Three-factor" (symmetric e-prop), "RF Three-factor" (RFLO) and "MDGL", respectively. We remark that the focus here is on comparing artificial to bio-plausible learning rules, rather than between biological rules. Finally, we note that tasks were learned with mostly comparable training accuracies for all learning rules, and that generalization gaps reflect a testing departure from these values. We refer the reader to the Appendix for more details (Appendix~\ref{scn:sim_details}). 
	
	\subsection{Generalization gap and loss landscape curvature}
	
	To study generalization performance, our first step is to compute the generalization gap empirically. Generalization gap is defined as train accuracy minus test accuracy; the larger the generalization gap, the worse the generalization performance. For various learning rules, we plot the generalization gap histogram at the end of training across runs with distinct initializations; different colors represent different learning rules (Figure~\ref{fig:Figure2}A-C). Notice that these bio-plausible rules achieve worse and more variable generalization performance than their machine learning counterpart (BPTT). 
	
	We now investigate if the generalization gap behavior described above correlates well with Loss landscape curvature. We use the leading eigenvalue of the loss' Hessian (where derivatives are taken with respect to model parameters) as a measure for curvature following previous practice~\cite{yao2018hessian} and note that it is practical for both empirical~\cite{yao2020pyhessian} and theoretical analyses (Theorem~\ref{thm:rank1_eval}). There exist other measures for flatness and due to the scale-dependence issue of Hessian spectrum~\cite{dinh2017sharp}, we also test using parameter scale-independent measures (see Appendix Figures~\ref{fig:EigSpectrum} and~\ref{fig:rel_flat}). 
	When we plot generalization gap points from Figure~\ref{fig:Figure2}A-C against the corresponding leading Hessian eigenvalue, a statistically significant correlation is observed (Figure~\ref{fig:Figure2}D-F). We also observed such correlation across runs with the learning rule fixed (Appendix Figure~\ref{fig:scatter_and_match}). We note that we did not expect this relationship to be very tight since in addition to the worse generalization gap on average, bio-plausible learning rules exhibit increased variance. This is an important and consistent trend we observed but might have been overlooked in previous studies.
	
	\begin{figure}[h!]
		\centering
		\includegraphics[width=0.99\textwidth]{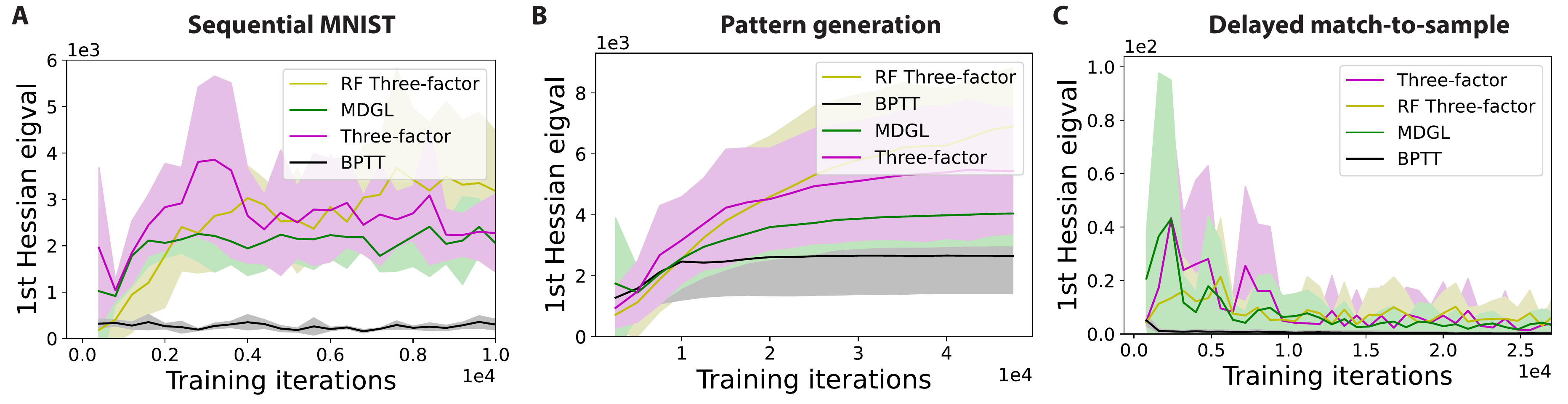}
		\caption{\textbf{Bio-plausible gradient approximations tend to approach high curvature regions in loss landscape}. Dominant Hessian eigenvalues are plotted throughout training for bio-plausible learning rules and BPTT. This analysis is done for A) sequential MNIST, B) pattern generation and C) delayed match-to-sample tasks. Solid lines/shaded regions: mean/standard deviation of dominant Hessian eigenvalue curves across five independent runs. 
		} 
		\label{fig:LamCurves}
	\end{figure}
	
	So far, we investigated the endpoints of optimization trajectories, where training performance has converged. Now, we visualize the whole training trajectory. This done is for two reasons: 1) account for early stopping that can halt training anywhere along the trajectory due to time constraints; 2) flatness during training could give indications of avoiding or escaping high-curvature regions. We observe that the biologically motivated gradient approximations tend to rapidly approach high-curvature regions compared to their machine learning counterpart (Figure~\ref{fig:LamCurves}). 
	Together, these results demonstrate a clear trend and a link between the generalization gap and loss landscape curvature: both being increased and more variable for bio-plausible rules. We also stopped BPTT early to match the test accuracy of the three-factor rule, and observed similar trends as in the main text (Appendix Table~\ref{table:1}). We also note that the curvature convergence behavior seems to be a shared problem of temporal truncations of the gradient (Appendix Figure~\ref{fig:LamCurves_trunc}), which is what the existing bio-plausible gradient approximations for RNNs are based on. 
	Next, we provide a theoretical argument as to why truncated temporal credit assignment rules favor high curvature regions of the loss landscape. 
	
	\subsection{Theoretical analysis: link between curvature and gradient approximation error}
	
\begin{figure}[t]
		\centering
		\includegraphics[width=0.99\textwidth]{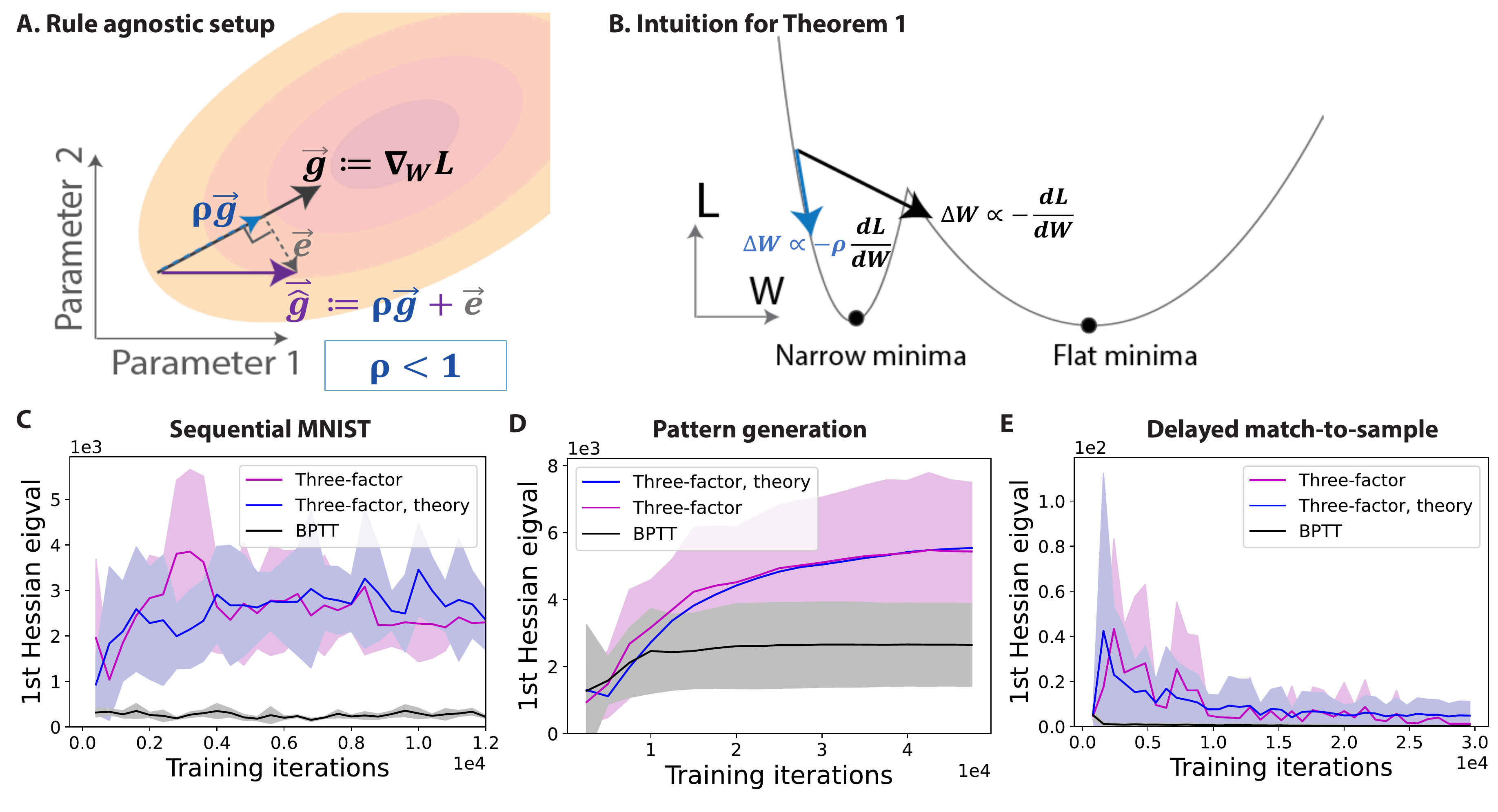}
		\caption{\textbf{Preference for high curvature regions connected to worse gradient alignment under certain conditions}. A) An arbitrary gradient approximation, $\vec{\hat{g}}$, its component along the gradient direction, $\rho \vec g$, and orthogonal to the gradient $\vec e$ (Eq.~\ref{eqn:rho_def}). The scalar $\rho$ represents the relative step length along the gradient direction. B) Illustration for Theorem~\ref{thm:rank1_eval}: if gradient $\vec g$ is aligned with the sharpest directions but error $\vec e$ is not (see Appendix Figure~\ref{fig:overlap}), smaller step length along the gradient direction can make it harder to step over narrow minima. C-E) Leading loss Hessian eigenvalue v.s. training iteration for the three-factor rule, BPTT, and modified BPTT (three-factor, theory). For the latter, step length along the gradient direction of BPTT was matched to three-factor rule by multiplying BPTT update with a factor of $\rho$, which recovers curvature trends of the three-factor rule.
		} 
		\label{fig:LamCurves_rholr}
	\end{figure}

	We discussed how generalization can be linked to curvature, and we now examine the link between curvature and gradient alignment during learning dynamics. We represent approximate gradients in a rule-agnostic manner, where an arbitrary approximation is represented in terms of its component along the gradient direction plus an arbitrary orthogonal vector (Figure~\ref{fig:LamCurves_rholr}A): 
	\begin{equation}
		\vec{\hat{g}} = \rho \vec g + \vec e, \label{eqn:rho_def}
	\end{equation}
	where $\vec g:=\nabla L(W_h) \in \mathbb{R}^{N^2}$ and $\vec{\hat{g}} \in \mathbb{R}^{N^2}$ are the exact and approximate gradients, respectively (we reshaped $\nabla L(W_h)$ into a vector here); $\vec e \in \mathbb{R}^{N^2}$ is an arbitrary orthogonal vector to $\vec g$. Here, scalar $\rho \in \mathbb{R}$ represents the relative step length along the gradient direction that the approximate rule is making. As we will see in Theorem~\ref{thm:rank1_eval}, $\rho$ is an important quantity in our analysis. One can easily compute $\rho$ from $\vec g$ and $\vec{\hat{g}}$ by $\rho=\frac{\vec{\hat{g}}^T \vec g}{\vec g^T \vec g}$. 
	
	We express weight updates as discrete dynamical systems (with weights $W_h$ as the state variables):
	\begin{align}
		W^{+}_h &\leftarrow W^{-}_h + F(W^{-}_{h}) = W^{-}_h -\eta \nabla L(W^{-}_h), \text{ for BPTT} \label{eqn:de_bptt} \\
		W^{+}_h &\leftarrow W^{-}_h + \hat{F}(W^{-}_h) = W^{-}_h -\eta \tilde \nabla L(W^{-}_h), \text{ for an approximate rule}  \label{eqn:de_eprop}, 
	\end{align}
	where $\eta$ is the learning rate, $\tilde \nabla$ is an approximate gradient, and $F : \mathbb{R}^{N^2} \rightarrow \mathbb{R}^{N^2}$ (resp. $\hat{F}: \mathbb{R}^{N^2}  \rightarrow \mathbb{R}^{N^2}$) denotes the map defined by BPTT (resp. an approximate) weight update rule. Notation $W^+$ and $W^-$ denotes $W$ at the next and current step, respectively. We note in passing that dynamical systems view of weight updates have been used previously~\cite{saxe2019mathematical,gidel2019implicit}.
	
	We introduce additional notations before presenting Theorem~\ref{thm:rank1_eval}. $J \in \mathbb{R}^{N^2 \times N^2}$ (resp. $\widehat{J} \in \mathbb{R}^{N^2 \times N^2}$) is the Jacobian of the dynamical system of BPTT (resp. an approximate rule) in Eq.~\ref{eqn:de_bptt} (resp. Eq.~\ref{eqn:de_eprop}). $\lambda^J_1 \in \mathbb{R}$ (resp. $\widehat{\lambda_1^J} \in \mathbb{R}$) is the leading eigenvalue for BPTT (resp. an approximate rule) Jacobian. $\lambda^H_1 \in \mathbb{R}$ is the leading eigenvalue of the loss' Hessian matrix. $W^*_B \in \mathbb{R}^{N \times N}$ (resp. $W^*_e \in \mathbb{R}^{N \times N}$) is the final fixed point for BPTT (resp. an approximate rule). We now present Theorem~\ref{thm:rank1_eval}.
	
	\begin{theorem}
		~\label{thm:rank1_eval}
		Consider an RNN defined in Eq.~\ref{eqn:vrnn} with a single scalar output $\hat{y}$ and least squares loss as in Eq.~\ref{eqn:loss_fcn0} presented only at the last time step $T$, and weights are updated according to the difference equation for BPTT~\ref{eqn:de_bptt} (resp. an approximate rule~\ref{eqn:de_eprop}) on a single example (stochastic gradient descent) using learning rate $\eta_B$ (resp. $\eta_e$). In the limit of stable fixed point convergence with zero training error, the dominant loss' Hessian eigenvalue attained by BPTT (resp. approximate rule) is bounded by $|\lambda^H_1(W^*_B)| < \frac{1}{\eta_B}$ (resp. $ |\lambda^H_1(W^*_e)| < \frac{1}{|\rho |\eta_e})$.
	\end{theorem}
	
	\begin{proof}
		Full proof is in Appendix~\ref{scn:proofs}. Here is a summary of the main steps involved: 
		\begin{enumerate}
			\item The Jacobian of BPTT dynamical system (Eq.~\ref{eqn:de_bptt}) is the loss' Hessian scaled by a constant; 
			\item Using the above relationship, we can bound $|\lambda^H_1|$ from $|\lambda^J_1|<1$, which is the condition for a discrete-time dynamical system to converge to a fixed point
			\item However, the link between the Jacobian of an approximate update rule and the loss' Hessian is less obvious. Thus, we derive a link between $|\widehat{\lambda_1^J}|$ and $|\lambda^J_1|$, and then apply the step above
		\end{enumerate}
	\end{proof}
	
	The consequence of the upper bound derived in Theorem~\ref{thm:rank1_eval} is that truncated gradient rules can converge to minima with a higher dominant Hessian eigenvalue than BPTT, with the leading eigenvalue bound inversely proportional to $|\rho|$ ($|\rho|<1$ usually). In practice, $\rho$ can vary depending on task settings and in our setup, we observed it to be somewhere between $0.02$ and $0.3$.
	We remark that this higher upper bound is consistent with the increased spread of curvature and generalization observed for bio-plausible rules in experiments.
	Theorem~\ref{thm:rank1_eval} highlights scalar $\rho$ (Eq.~\ref{eqn:rho_def}), relative step length along the gradient direction, as an important factor in the curvature bound. To test that, we eliminate the factor of $\rho$ by reducing the learning rule of BPTT such that its step length is matched to that of the three-factor rule. This resulted in the blue curves in Figure~\ref{fig:LamCurves_rholr}, which is still trained using BPTT but with the update scaled by a factor of $\rho$. By matching the step length of BPTT and a three-factor rule along the gradient direction, similar convergence behaviors were observed. Similar observations were also made when the matching step experiment was repeated at three times the learning rate for all rules (Appendix Figure~\ref{fig:scatter_and_match}C). {\bf This result then attributes the curvature preference behavior to relative step length along the gradient direction, and thereby indicating a link between curvature and gradient alignment under certain conditions}. Consistent with earlier results, when the step length of BPTT is matched to that of a three-factor rule, its generalization performance also worsened (Appendix Figure~\ref{fig:gengap_rholr}). 
	
	We make a few more remarks regarding Theorem~\ref{thm:rank1_eval}. $\rho$ is usually less than $1$ because otherwise the additional orthogonal component $\vec e$ would imply a larger update size for the approximate rule compared to BPTT. This would make the approximate rule more prone to numerical instabilities (Eq.~\ref{eqn:wnorm}) and would require a change in solver hyperparameters such as overall learning rate which in turn, influence the update size. In our experiments, when the three-factor rule was scaled up to match the along-gradient update sizes of BPTT, we quickly ran into numerical overflow (values of NaNs in the network), which is expected for a very large learning rate. The exact point for when this numerical instability is reached depends on many factors such as the model, task, numerical precision as well as the consistency of update direction. We have also equated numerical instabilities in simulations as a proxy for situations problematic for the brain in Discussion. Curiously, $\vec e$ did not seem to play a helpful role in finding flatter minima; this could be due to that approximation error $\vec e$ is not well aligned with the sharpest directions (Appendix Figure~\ref{fig:overlap}). In fact, $\vec e$ being orthogonal to the leading Hessian eigenvectors is a consequence of the assumptions behind Theorem~\ref{thm:rank1_eval}. Despite making assumptions including scalar output, MSE loss and loss available at the last step, our empirical results indicate that our conclusions also extend to other setups: vector output, cross-entropy loss and loss that accumulates over time steps (Figure~\ref{fig:LamCurves_rholr}). In Appendix~\ref{scn:thm_gen}, we discuss the generality of Theorem~\ref{thm:rank1_eval} by using quadratic expansion of the loss function and assuming that $\vec e$ is orthogonal to the leading Hessian eigenvectors. 
	
	One might assume increasing the learning rate of the approximate rules could compensate for the reduced step length due to $\rho$. However, this can raise other issues. Suppose $\Delta W = -\eta \vec g$ for the exact gradient and we increase the learning rate of the approximate update until $\widehat{\Delta W} = -\eta \vec g - \eta \vec e$ (magnitude of $ \vec e$ is scaled accordingly), and because $\vec g \perp \vec e$:  
	\begin{equation}
		\| \widehat{\Delta W} \|^2 = \eta^2 \|\vec g\|^2 + \eta^2 \|\vec e\|^2 > \eta^2 \|g\|^2 = \| \Delta W \|^2 \label{eqn:wnorm}
	\end{equation}
	In other words, if approximate updates $\widehat{\Delta W}$ make the same amount of progress as BPTT along the gradient direction, $\widehat{\Delta W}$ would have a larger magnitude due to the orthogonal component $\vec e$, and large update magnitude can be very problematic for numerical stability~\cite{goodfellow2016deep}. Thus, the magnitude of $\vec e$ limits the learning rate that can be used. \textbf{Because of the numerical issues associated with increasing the learning rate for approximate rules (due to $\vec e$), the differences in generalization and curvature convergence between rules cannot be reduced by increasing the learning rate for approximate rules.} To balance between this numerical stability issue and the potential benefit of large learning rates, one can consider using a large learning rate early in training to prevent premature stabilization in sharp minima followed by gradual decay to mitigate the stability issue (see Appendix Figure~\ref{fig:LamCurves_lrdecay}). 
	
	Taken together, Theorem~\ref{thm:rank1_eval} and Eq.~\ref{eqn:wnorm} connect gradient alignment with the curvature of the converged solution under certain conditions. \textbf{For an approximation that is aligned poorly with the exact gradient, large orthogonal approximation error vector $\vec e$ limit the step length for numerical stability reasons (Eq.~\ref{eqn:wnorm}) and small relative step length $\rho$ correspond to a larger curvature bound (Theorem~\ref{thm:rank1_eval}). }
	
	\section{Conclusion}
	
	While developing bio-plausible learning rules is of interest for both answering neuroscience questions and searching for more efficient training strategies for artificial networks, the generalization properties of solutions found by these rules are severely underexamined. 
	Through various well-known machine learning and working memory tasks, we first demonstrate empirically that existing bio-plausible temporal credit assignment rules attain worse generalization performance, which is consistent with their tendency to converge to high-curvature regions in loss landscape. Second, our theoretical analysis offers an explanation for this preference for high curvature regions based on worse alignment to the true gradient. This regime corresponds to the situation where the step length along the gradient direction is small and the approximation error vector is large. Finally, we test this theory empirically by matching the relative step length along the gradient direction resulting in similar convergence behavior (Figure~\ref{fig:LamCurves_rholr}).

	\section{Discussion}
	
	Our study --- a stepping stone toward understanding biological generalization using deep learning methods --- raises many exciting questions for future investigations, both on the front of stronger deep learning theory and more sophisticated biological ingredients.
	
	\paragraph{Deep learning theory and its implications:}
	In this study, we investigate generalization properties using loss landscape flatness, a promising predictor with recent rigorous connection to generalization gap~\cite{petzka2021relative}. Yet, to what extent can flatness explain generalization is still an open question in deep learning. For instance, curvature is a local measure, which means that its informativeness of robustness against global perturbation is limited. Moreover, the theoretical association between flatness and generalization is provided in the form of upper bound~\cite{petzka2021relative,tsuzuku2020normalized} and the bound may not always be tight, which is consistent with more variability in the generalization gap for bio-plausible learning rules but offers less predictive power. Despite observing a significant correlation between the generalization gap and a curvature-based measure in Figure~\ref{fig:Figure2}, the relationship appears to be messy, suggesting other factors involved in explaining generalization. Given that developing better predictors of generalization is still a work in progress~\cite{jiang2020neurips}, we anticipate stronger theoretical tools to be applied for studying biological generalization in the future.
	Our results are also consistent with existing findings linking learning rate to loss landscape curvature in deep networks~\cite{jastrzkebski2018finding}. In the case of bio-plausible gradient approximations, small step length along the gradient direction cannot be compensated by increasing the learning rate, as that would inadvertently increase the error vector, causing numerical issues (Eq.~\ref{eqn:wnorm}). Curiously, the approximation error vector $\vec e$ did not seem to play a role other than restricting the learning rate. While it is well-known that stochastic gradient noise (SGN) can help with finding flat minima due to the alignment of SGN covariance and Hessian near minima~\cite{zhu2018anisotropic,wei2019noise,xie2020diffusion,smith2017bayesian}, that may not apply to approximation error vector $\vec e$ resulting primarily from temporal truncation of the gradient (see Figure~\ref{fig:Figure1}A and Appendix Figure~\ref{fig:overlap}). This indicates that noise with different properties (e.g. different directions) could affect generalization differently, thereby motivating future investigations into how a broad range of biological noises --- which may differ from noise in ML optimization (e.g. SGN) --- can impact generalization. 
	
	Moreover, our results are closely related to a series of studies that examined the "catapulting" behavior in learning~\cite{cohen2021gradient,lewkowycz2020large,jastrzebski2020break,gilmer2021loss,arora2022understanding}. This can happen when the second order Taylor term of the loss function would dominate over the first, which would cause learning to cross the threshold for step size stability and "catapult" into a flatter region that can accommodate the step size. If the truncation noise is only aligned with the eigendirections associated with negligible eigenvalues, then it can only have limited contributions to the second-order Taylor term. On top of that, the orthogonal noise term would require a smaller learning rate to be used to avoid numerical issues, as explained earlier. Overall, these would lead to a weaker second-order Taylor term relative to the first for bio-plausible temporal credit assignment rules, which would then increase the threshold for step size stability. This increased stability is closely tied to the greater dynamical stability (for the weight update difference equation) of approximation rules predicted by Theorem~\ref{thm:rank1_eval}, due to the correspondence between loss' Hessian matrix and the Jacobian matrix of the weight update difference equation (the correspondence is explained in Theorem~\ref{thm:rank1_eval} proof in Appendix~\ref{scn:proofs}).    
	
	\paragraph{Toward more detailed biological mechanisms:}
	On the front of more sophisticated biological ingredients that may improve generalization performance, we see two lines of approaches: {\bf 1-} develop bio-plausible learning rules that align better with the gradient, as suggested by our rule-agnostic analysis (Theorem~\ref{thm:rank1_eval}, Figure~\ref{fig:LamCurves_rholr}); and {\bf 2-} instead of studying learning rules in isolation, consider also other neural circuit components~\cite{ivanov2021increasing,braun2021online,dapello2021neural,de2016astrocytes,murray2020remembrance,zador2019critique,kudithipudi2022biological,billeh2020systematic} that could interact with the learning rule. An important component would be the architecture, including connectivity structure and neuron model, found through evolution~\cite{zador2019critique} (see also~\cite{geadah2022goal}). To address our main question, this study varies learning rules while holding data, objective function and architecture constant (see~\cite{richards2019deep}). However, these different components can interact, and more sophisticated architecture can facilitate task learning~\cite{jones2020can,perez2021neural,logiaco2021thalamic,yang2021next,teeter2018generalized,winston2022heterogeneity,geadah2022goal,stockl2021probabilistic}. Given the exploding parameter space resulting from such interactions, we believe it requires careful future analysis and is outside of the scope for this one paper. 
	
	Additionally, learning rate modulation~\cite{li2019towards,maheswaranathan2021reverse} could be one of many possible remedies employed by the brain. We conjecture that neuromodulatory mechanisms could be coupled with these learning rules to improve the convergence behavior through our scheduled learning rate experiments (Appendix Figure~\ref{fig:LamCurves_lrdecay}), where an initial high learning rate could prevent the learning trajectory from settling in sharp minima prematurely followed by gradual decay to avoid instabilities. One possible way to realize such learning rate modulation could be through serotonin neurons via uncertainty tracking, where the learning rate is high when the reward prediction error is high (this can happen at the beginning of learning)~\cite{grossman2021serotonin}. Since the authors of~\cite{grossman2021serotonin} showed that inhibiting serotonin led to failure in learning rate modulation, we conjecture that such inhibition might have an impact on the generalization performance of learning outcomes. On the topic of balancing numerical instabilities and potential advantages of large learning rates, while the analog nature of biology may seem to avoid finite precision representation in digital computers that give rise to numerical instabilities, the same problems that lead to numerical instabilities in digital computers, such as big ranges between quantities added or multiplied, remains an issue for biology since quantities must be stored in noisy activity patterns of neurotransmitter release. Future investigations could investigate potential homeostatic mechanisms that regulate biological quantities to avoid such instabilities, thereby enabling larger "learning rates” to be used so as to find flatter minima. Other ingredients for future investigations could include intrinsic noise with certain structures~\cite{orvieto2022anticorrelated,zhou2019toward,dapello2021neural} (including directions and bias/variance properties) that would make them more favorable for generalization. Taken together, we hope to see follow-up investigations --- riding on the rapid advancements both at the front of deep learning theory and sophisticated biological mechanisms --- into how the brain attains solutions that generalize. 
	
	\section{Acknowledgement}
	We thank Gauthier Gidel, Jonathan Cornford, Aristide Baratin, Thomas George and Mohammad Pezeshki for insightful discussions and helpful feedback at an early stage of this project. We also thank Henning Petzka and Michael Kamp for helpful email exchanges. This research was supported by NSERC PGS-D (Y.H.L.); NSF AccelNet IN-BIC program, Grant No. OISE-2019976 AM02 (Y.H.L.); Vanier Canada Graduate scholarship (A.G.); Healthy Brains for Healthy Lives fellowship (A.G.); CIFAR Learning in Machines and Brains Program (B.A.R.), NSERC Discovery Grant RGPIN-2018-04821 (G.L), FRQS Research Scholar Award, Junior 1 LAJGU0401-253188 (G.L.), Canada Research Chair in Neural Computations and Interfacing (G.L.), Canada CIFAR AI Chair program (G.L. \& B.A.R.). We also thank the Allen Institute founder, Paul G. Allen, for his vision, encouragement, and support. 
	
	\bibliographystyle{unsrt}
	\bibliography{ref_main}
	
	\section*{Checklist}

	The checklist follows the references.  Please
	read the checklist guidelines carefully for information on how to answer these
	questions.  For each question, change the default \answerTODO{} to \answerYes{},
	\answerNo{}, or \answerNA{}.  You are strongly encouraged to include a {\bf
		justification to your answer}, either by referencing the appropriate section of
	your paper or providing a brief inline description.  For example:
	\begin{itemize}
		\item Did you include the license to the code and datasets? \answerYes{See Section~\ref{gen_inst}.}
		\item Did you include the license to the code and datasets? \answerNo{The code and the data are proprietary.}
		\item Did you include the license to the code and datasets? \answerNA{}
	\end{itemize}
	Please do not modify the questions and only use the provided macros for your
	answers.  Note that the Checklist section does not count towards the page
	limit.  In your paper, please delete this instructions block and only keep the
	Checklist section heading above along with the questions/answers below.

	\begin{enumerate}

		\item For all authors...
		\begin{enumerate}
			\item Do the main claims made in the abstract and introduction accurately reflect the paper's contributions and scope?
			\answerYes{To make this easier for the readers, we have referred to the pertinent Figures and Theorem to support our summary of contributions in Introduction.}
			\item Did you describe the limitations of your work?
			\answerYes{Limitations and future work can be found in our Discussion section.}
			\item Did you discuss any potential negative societal impacts of your work?
			\answerNo{This work contributes to the theoretical understanding of biologically-plausible learning rules for recurrent neural networks. As such, we do not anticipate any direct ethical or societal impact. In the long-term, our work can have impact on related research communities such as neuroscience and deep learning, with societal impact depending on the development of these fields. }
			\item Have you read the ethics review guidelines and ensured that your paper conforms to them?
			\answerYes{We have carefully read the ethics review guidelines and confirm that this
				paper conforms to them.}
		\end{enumerate}
		
		\item If you are including theoretical results...
		\begin{enumerate}
			\item Did you state the full set of assumptions of all theoretical results?
			\answerYes{We state all assumptions precisely when we first presented the Theorem in Results, We further details, discussions and proofs in the appendix. }
			\item Did you include complete proofs of all theoretical results?
			\answerYes{Proofs can be found in Appendix~\ref{scn:proofs}.}
		\end{enumerate}
		
		\item If you ran experiments...
		\begin{enumerate}
			\item Did you include the code, data, and instructions needed to reproduce the main experimental results (either in the supplemental material or as a URL)?
			\answerYes{Anonymized code link is provided in Appendix~\ref{scn:sim_details}.}
			\item Did you specify all the training details (e.g., data splits, hyperparameters, how they were chosen)?
			\answerYes{All training details are provided in Appendix~\ref{scn:sim_details}. }
			\item Did you report error bars (e.g., with respect to the random seed after running experiments multiple times)?
			\answerYes{We tried to provide this information in all applicable figures. This is stated as "Solid lines/shaded regions: mean/standard deviation obtained at the end of training across five independent runs" in the legends of several figures.}
			\item Did you include the total amount of compute and the type of resources used (e.g., type of GPUs, internal cluster, or cloud provider)?
			\answerYes{Information pertaining to computing resources and simulation time can be found in Appendix~\ref{scn:sim_details}.}
		\end{enumerate}
		
		\item If you are using existing assets (e.g., code, data, models) or curating/releasing new assets...
		\begin{enumerate}
			\item If your work uses existing assets, did you cite the creators?
			\answerYes{For dataset, we cited MNIST as well as representative neuroscience papers for the two neuroscience motivated tasks. We have also cited existing coding assets in Appendix~\ref{scn:sim_details}. }
			\item Did you mention the license of the assets?
			\answerYes{We mentioned that in Appendix~\ref{scn:sim_details}.}
			\item Did you include any new assets either in the supplemental material or as a URL?
			\answerNo{}
			\item Did you discuss whether and how consent was obtained from people whose data you're using/curating?
			\answerNA{}
			\item Did you discuss whether the data you are using/curating contains personally identifiable information or offensive content?
			\answerNA{}
		\end{enumerate}

		\item If you used crowdsourcing or conducted research with human subjects...
		\begin{enumerate}
			\item Did you include the full text of instructions given to participants and screenshots, if applicable?
			\answerNA{}
			\item Did you describe any potential participant risks, with links to Institutional Review Board (IRB) approvals, if applicable?
			\answerNA{}
			\item Did you include the estimated hourly wage paid to participants and the total amount spent on participant compensation?
			\answerNA{}
		\end{enumerate}

	\end{enumerate}
	
	
	\newpage
	
	\appendix
	
	\section{Methods} \label{scn:methods}
	
	\subsection{Hessian eigenspectrum analysis}
	
	As mentioned, we focus on the leading Hessian eigenvalue because it has been used previously~\cite{yao2018hessian,jastrzkebski2018relation} and is very feasible for both empirical~\cite{yao2020pyhessian,sagun2016eigenvalues} and theoretical analyses (Theorem~\ref{thm:rank1_eval}). 
	The leading Hessian eigenvalue can be computed by performing power iterations on the Hessian vector product without knowing the full Hessian matrix (Algorithm 2 in~\cite{yao2020pyhessian}). To find multiple top eigenvalues, e.g. top 200 eigenvalues, one can use the generalized power method via QR decomposition~\cite{trefethen1997numerical}. We focused on the loss' Hessian for recurrent weights but observed a similar trend for input weights as well. We set the tolerance for stopping to $1e-6$. 
	
	Due to the scale-dependence issue of Hessian spectrum~\cite{dinh2017sharp}, we also used scale-independent measures. For instance, we examined the power-law decay coefficient for the Hessian eigenvalues (Figure~\ref{fig:EigSpectrum} in Appendix). We also looked at the recently proposed relative flatness measure~\cite{petzka2021relative} (Figure~\ref{fig:rel_flat} in Appendix). We used the code in~\cite{stringer2019high} to fit a power-law distribution to the top 200 eigenvalues. We found similar results had we chosen the top 50 or 100 eigenvalues instead, and 200 was chosen mainly due to computational load. We note that the link of power-law decay to generalization has also been examined in some recent studies~\cite{ghosh2022investigating,xie2022power,chen2021analysis}. 
	
	\subsection{Network setup and learning rule implementations} \label{scn:net_details}
	
	\textbf{Neuron Model}: We consider a discrete-time implementation of a rate-based recurrent neural network (RNN) similar to the form in~\cite{ehrlich2021psychrnn}. The model denotes the internal hidden state as $h_t$ and the observable states, i.e. firing rates, as $f(h_{t})$ at time $t$, and we use ReLU activation for $f$. The dynamics of those states are governed by 
	\begin{align}
		h_{t+1} &= \alpha h_t + (1-\alpha) \left( W_h f(h_t) + W_x x_t \right), \label{eq_rate1}
	\end{align}
	where $\alpha=e^{-dt/\tau_m}$ denotes the leak factor for simulation time step $dt$ and membrane time constant $\tau_m$, $W_h$ denotes the weight of the recurrent synaptic connection, $W_x$ denotes the strength of the input synaptic connection and $x_t$ denotes the external input at time $t$. we use subscripts to represent indices of neurons and time steps. For instance, $h_{i,t}$ represents the hidden activity $h$ of neuron $i$ at time $t$. $W_{h,ij}$ represents the $(ij)^{th}$ entry of recurrent weight matrix $W_h$. Model in Eq.~\ref{eq_rate1} was used for the sequential MNIST and pattern generation tasks. 
	
	We mention in passing that the choice of ReLU activation, which has a discontinuous first derivative, means that the loss Hessian matrix is not guaranteed to be symmetric. A real matrix that is not symmetric can have complex eigenvalues come in conjugate pairs, and if they were amongst the top eigenvalues, power iterations may not converge. However, all iterations have converged in our experiments as mentioned above. Also, because of potential technical issues resulting from non-symmetric Hessian matrices, we foresee challenges in applying our methodology to spiking neural networks (SNNs), which have discontinuous activation functions. Due to the energy efficiency and biological realism of SNNs~\cite{neftci2019surrogate,roy2019towards,cramer2022surrogate,huh2018gradient,zenke2018superspike,zhang2020temporal,xiao2021training,perez2021sparse,li2021differentiable,skatchkovsky2021learning,zenke2020spike}, we believe extending to SNNs is an important future direction.
	
	For the delayed match to sample task, which is a working memory task, it was found in~\cite{liu2021pnas} and~\cite{Bellec2020} that units with an adaptive threshold as an additional hidden variable can play an important role in the computing capabilities of RNNs. Thus, we implemented adaptive threshold neuron units~\cite{teeter2018generalized} for that task. In our rate-based implementation, this turns out to be a simple addition of a second hidden variable $b_t$ that represents the dynamic threshold component:
	\begin{align}
		h_{t+1} &= \alpha h_t + (1-\alpha) \left( W_h f(h_t - b_t) + W_x x_t \right),  \cr
		b_{t+1} &= \beta b_{t} + (1-\beta) f(h_t - b_t), \label{eq_arate1}
	\end{align}
	where $b_{j,t}$ denotes the dynamic threshold that adapts based on past neuron activity. The decay factor $\beta$ is given by $e^{-dt/\tau_b}$ for simulation time step $dt$ and adaptation time constant $\tau_b$, which is typically chosen on the behavioral task time scale~\cite{liu2021pnas}. 
	
	\textbf{Network output and loss function}:  
	
	Readout $\hat y$ is defined as
	\begin{equation}
		\hat y = \langle w, f(h_t)\rangle
	\end{equation}
	for readout weights $w$.
	
	We quantify how well the network output $\hat y$ matches the desired target $y$ using loss function $L$, which is defined as
	\begin{equation}
		L(W_h) = \begin{cases}
			\frac{1}{2T B}\sum_{i=1}^B \sum_{t=1}^T \sum_{k=1}^{N_{out}} (\hat y^{(i)}_{k,t} - y^{(i)}_{k,t})^2, \text{ for regression tasks} \cr
			\frac{-1}{T B} \sum_{i=1}^B \sum_{t=1}^T \sum_{k=1}^{N_{out}} \pi^{(i)}_{k,t} log \hat \pi^{(i)}_{k,t}, \text{ for classification tasks} 
		\end{cases} \label{eqn:loss_fcn}
	\end{equation}
	for target output $y$, task duration $T$, $N_{out}$ output neurons and batch size $B$. $\pi_{k,t}$ is the one-hot encoded target and $\hat \pi_{k,t} = \text{softmax}_k (\hat y_{1,t}, \dots, \hat y_{N_{OUT},t}) = \exp(\hat y_{k,t})/\sum_{k'} \exp(\hat y_{k',t})$ is the predicted category probability. 
	
	\textbf{Biological gradient approximations (truncation-based)} 
	
	The goal of this subsection is to explain where the approximation happens for each of the bio-plausible learning rules. For full details regarding these rules, we encourage the reader to refer to the respective references. We start by writing down the gradient in terms of real-time recurrent learning (RTRL) factorization: 
	\begin{eqnarray}
		\frac{\partial L}{\partial W_{h, ij}} &=&\sum_{l,t}\frac{\partial L}{\partial h_{l,t}} \frac{\partial h_{l,t}}{\partial W_{h, ij}}, \label{eqn:sum_tl} 
	\end{eqnarray}
	
	Key problems that RTRL poses to biological plausibility and computational cost reside in the second factor $\frac{\partial h_{l,t}}{\partial W_{h, ij}}$ that arises during the factorization of the gradient (Eq.~\ref{eqn:sum_tl}). The factor $\frac{\partial h_{l,t}}{\partial W_{h, ij}}$ keeps track of all recursive dependencies of $h_{l,t}$ on weight $W_{h,ij}$ arising from recurrent connections. These recurrent dependencies can be obtained recursively as follows: 
	\begin{align}
		\frac{\partial h_{l,t}}{\partial W_{h, ij}} &= \frac{\partial h_{j,t}}{\partial W_{h,ij}} + \sum_m \frac{\partial h_{l,t}}{\partial h_{m,t-1}} \frac{\partial h_{m,t-1}}{\partial W_{h,ij}} \cr 
		&= \frac{\partial h_{l,t}}{\partial W_{h,ij}} + \frac{\partial h_{l,t}}{\partial h_{l,t-1}} \frac{\partial h_{l,t-1}}{\partial W_{h,ij}} + 
		{\underbrace{\textstyle \sum_{m\neq l} W_{h,lm} f'(h_{m,t-1}) \frac{\partial h_{m,t-1}}{\partial W_{h,ij}} }_{\mathclap{ \text{\normalsize depends on all weights $W_{h,lm}$}}}}. \label{eqn:s_triple}
	\end{align}
	Thus, the factor $\frac{\partial h_{l,t}}{\partial W_{h, ij}}$ poses a serious problem for biological plausibility: it involves \textbf{nonlocal} terms that should be inaccessible to neural circuits, i.e. that knowledge of all other weights in the network is required in order to update the weight $W_{h,ij}$. 
	
	\textbf{RFLO}~\cite{murray2019local} (labeled as "RF Three-factor") and \textbf{symmetric e-prop}~\cite{Bellec2020} (labeled as "Three-factor") seek to address this by truncating the expensive nonlocal terms in Eq.~\ref{eqn:s_triple} so that the updates to weight $W_{h,ij}$ would only depend on pre- and post-synaptic activity:
	\begin{align}
		\widehat{\frac{\partial h_{l,t}}{\partial W_{h, ij}}} &= \begin{cases} 
			\frac{\partial h_{i,t}}{\partial W_{h,ij}} + \frac{\partial h_{i,t}}{\partial h_{i,t-1}} \widehat{\frac{\partial h_{i,t-1}}{\partial W_{h,ij}}},  & l = i \\
			0, & l \neq i
		\end{cases}
	\end{align}
	which results in a much simpler factor than the triple tensor in Eq.~\ref{eqn:s_triple}.
	
	After the truncation, RFLO and e-prop implement:
	\begin{eqnarray}
		\widehat{\frac{\partial L}{\partial W_{h, ij}}} &=& \sum_{t}\frac{\partial L}{\partial h_{i,t}} \widehat{\frac{\partial h_{i,t}}{\partial W_{h, ij}}}, \\ 
		\widehat{\frac{\partial h_{i,t}}{\partial W_{h, ij}}} &=& 
		\frac{\partial h_{i,t}}{\partial W_{h,ij}} + \frac{\partial h_{i,t}}{\partial h_{i,t-1}} \widehat{\frac{\partial h_{i,t-1}}{\partial W_{h,ij}}}.
	\end{eqnarray}
	
	The main difference between symmetric e-prop and RFLO implementation is that symmetric feedback is used for symmetric e-prop, i.e. output weight $w$ is used as the feedback weight for the $\frac{\partial E}{\partial h}$, whereas RFLO uses fixed random feedback weights~\cite{lillicrap2016random} for greater biological plausibility. We note in passing that the authors of e-prop have tested their formulation with fixed random feedback weights as well. 
	\textbf{MDGL}~\cite{liu2021pnas} also truncates RTRL, but it restores some of the non-local dependencies -- those within one connection step --- that could potentially be communicated via mechanisms similar to the abundant cell-type-specific local modulatory signaling unveiled by recent transcriptomics data~\cite{smith2020new,smith2019single}. With that, the expensive memory trace term in Eq.~\ref{eqn:s_triple} becomes
	\begin{equation}
		\frac{\partial h_{l,t}}{\partial W_{h, ij}} \approx
		\begin{cases}
			W_{h,li} f'(h_{i,t-1}) \widehat{\frac{\partial h_{i,t-1}}{\partial W_{h,ij}}}, & i\neq l \vspace{0.1in} \\
			\frac{\partial h_{i,t}}{\partial W_{h,ij}} + \frac{\partial h_{i,t}}{\partial h_{i,t-1}} \widehat{\frac{\partial h_{i,t-1}}{\partial w_{ij}}}, & i=l
		\end{cases}
		\label{tensor_approx}
	\end{equation}
	
	MDGL involves one additional approximation: replace $W_{h,li}$ with type-specific weights $W_{ab}$ to mimic the cell-type-specific nature of local modulatory signaling (for cell $i$ in group $a$ and cell $j$ in group $b$, where $a, b \in C$ for a total of $C$ cell groups). For simplicity, we just used $W_{ab}=W_{li}$, i.e. without cell-type approximation. This results in overall MDGL implementation as 
	\begin{eqnarray}
		\widehat{\frac{\partial L}{\partial W_{h, ij}}} &=& \sum_{t}\frac{\partial L}{\partial h_{i,t}} \widehat{\frac{\partial h_{i,t}}{\partial W_{h, ij}}} + \widehat{\frac{\partial h_{i,t-1}}{\partial W_{h, ij}}} \sum_{l} W_{h,li} \frac{\partial L}{\partial h_{l,t}},  
	\end{eqnarray} 
	Interpretation of the above update rule in terms of biological processes can be found in the MDGL paper~\cite{liu2021pnas,liu2021solution}. 
	
	We note that input, recurrent and output weights were all being trained. This section illustrates the approximate gradient for updating recurrent weights $W_h$, and similar expressions were obtained for updating input weights $W_x$. The approximations, however, did not apply to output weights, as the gradient for that would not violate the aforementioned issue of nonlocality (Eq.~\ref{eqn:s_triple}). 
	
	\subsection{Simulation details} \label{scn:sim_details}
	
	We used TensorFlow~\cite{abadi2016tensorflow} version 1.14 and based it on top of~\cite{bellec2018long}. We modified the code for rate-based neurons (Eq.~\ref{eq_rate1} and~\ref{eq_arate1}). \footnote{Our code is available: \url{https://github.com/Helena-Yuhan-Liu/BiolHessRNN}.} We used the code in~\cite{stringer2019high} for the power-law analysis (Figure~\ref{fig:EigSpectrum} in Appendix). SGD optimizer was used to study the effect of gradient approximation in isolation without the complication of additional factors, as Adam optimizer with adaptive learning rate could convolute our matching step length experiments in Figure~\ref{fig:LamCurves_rholr}. That said, we verified that the curvature convergence behavior is also observed for Adam optimizer (Figure~\ref{fig:LamCurves_adam} in Appendix. Learning rates were optimized by picking within $\{3e-4, 1e-3, 3e-3, 1e-2, 3e-2, 1e-1 \}$ for each algorithm. For the sequential MNIST task, we explored batch sizes within $\{64, 256, 1024 \}$. For the sequential MNIST task, these hyperparameters were optimized based on validation performance (the validation set loaded using $tensorflow.examples.tutorials.mnist$). For the two other tasks, these hyperparameters were optimized based on the training performance, but we also tried optimizing on the test set and observed similar trends. Trainings were stopped when both the loss and leading Hessian eigenvalue stabilized. As stated, we repeated runs with different random initialization to quantify uncertainty and weights were initialized similarly as in~\cite{murray2019local}.
	
	Simulations were completed on a computer server with x2 20-core Intel(R) Xeon(R) CPU E5-2698 v4 at 2.20GHz. The average time to complete one run of sequential MNIST, pattern generation and delayed match to sample tasks in Figure~\ref{fig:LamCurves} were approximately 2 hours, 1 hour and 1 hour, respectively. Since the computation of second order gradient becomes prohibitively expensive as sequence length $T$ becomes large, all tasks involved no more than 50 time steps. For instance, this was achieved for the sequential MNIST task using the row-by-row implementation. Using fewer time steps, however, should not affect the general trend as the gradient truncation effects were still significant. Because of the use of fewer steps, we dropped the leak factor $\alpha$ in Eq.~\ref{eq_rate1} (i.e. set $\alpha=0$). 
	
	For the matching step length experiments (Figure~\ref{fig:LamCurves_rholr}), we simply obtained $\rho = \frac{\vec{\hat{g}}^T \vec g}{\vec g^T \vec g}$ for the three-factor learning rule and scaled BPTT updates by that amount. For scheduled learning rate experiments (Figure~\ref{fig:LamCurves_lrdecay}), the additional hyperparameters included initial learning rate, decay percentage and decay frequency. We used an initial learning rate that was three times the uniform rate (used in other figures) and decay the learning rate by $80\%$ every X iterations, where X was roughly the total number of training iterations (used in other figures) divided by 30. Since the point of that figure was to show that learning rate scheduling could lead to flatter minima than using a fixed learning rate, we did not search extensively across these additional hyperparameters as the first set of hyperparameters we tried was enough to demonstrate that point. 
	
	For the pattern generation task, our network consisted of $N=30$ neurons described in Eq.~\ref{eq_rate1}. Input to this network was provided by a random Gaussian input ($N_{in}=1$). The fixed target signal had a duration of 50 steps and was given by the sum of four sinusoids, with a fixed period of 10, 40, 70 and 100 steps. For learning, we used the mean squared loss function. Training for this task used full batch. For testing, we perturbed the input with additive zero-mean Gaussian noise (with $\sigma$ picked uniformly between $0$ and $0.2$ across runs), to mimic the situation where the agent had to faithfully produce the desired pattern even under perturbations. Unlike the other tasks, this task measures accuracy by mean squared error, for which the lower the better. To maintain the convention of a higher generalization gap being worse, the generalization gap for this task was computed by test error minus train error. 
	
	For the delayed match to sample task, our network consisted of $N=100$ neurons, which include 50 neurons with (Eq.~\ref{eq_arate1}) and 50 neurons without (Eq.~\ref{eq_rate1}) threshold adaptation. The task involved the presentation of two sequential cues, each taking on a binary value, lasting 2 steps and separated by a delay of 16 steps. Input to this network was provided by $N_{in}=2$ neurons. The first (resp. second) input neuron sent a value of 1 when the presented cue took on a value of 1 (resp. cue 0), and 0 otherwise. The network was trained to output 1 (resp. 0) when the two cues have matching (resp. non-matching) values. For learning, we used the cross-entropy loss function and the target corresponding to the correct output was given at the end of the trial. Training for this task used full batch. For testing, we tested on increased delay, with the period picked uniformly between the training delay period and twice the training delay period, to mimic situations where the animal has to hold the memory longer than it did during the learning phase. 
	
	For the sequential MNIST task~\cite{lecun1998mnist}, our network consisted of $N=128$ neurons described in Eq.~\ref{eq_rate1}. Input to this network was provided by $N_{in}=28$ units that represented the grey-scaled value of a single row, totaling 28 steps and the network prediction was made at the last step. For learning, we used the cross-entropy loss function and the target corresponding to the correct output was given at the end of the trial. For testing, we used the existing MNIST test set~\cite{lecun1998mnist} with additive zero-mean Gaussian input noise. As mentioned, this task did not train with a full batch but we found the trend to hold across different batch sizes.
	
	Finally, we note that comparisons between BPTT and approximate rules were done at comparable training accuracies for the pattern generation and delayed match to sample tasks. For the sequential MNIST task, the three-factor rule achieved only around $70\%$ training accuracy, but the training accuracy did not explain the curvature convergence behavior. To see this, while three-factor theory (blue in Fig 4), which corresponds to BPTT with reduced step length, achieves an accuracy of $>95\%$ but still attains similar curvature convergence to that of the three-factor rule.
	
	\newpage 
	
	\section{Theorem~\ref{thm:rank1_eval}} \label{scn:proofs}
	
	\subsection{Proof of Theorem~\ref{thm:rank1_eval}}
	
	\begin{proof}
		
		First, we note that the Jacobian of the dynamical system for BPTT update (Eq.~\ref{eqn:de_bptt}) is simply the loss Hessian scaled by $-\eta$. This implies that
		\begin{align}
			|\lambda^J_1| &= \eta_B |\lambda^H_1| \label{eqn:Jlam_bptt},
		\end{align}
		where we remind the reader that $\lambda^J_1 \in \mathbb{R}$ (resp. $\widehat{\lambda_1^J} \in \mathbb{R}$) is the leading eigenvalue for BPTT (resp. an approximate rule) Jacobian; $\lambda^H_1 \in \mathbb{R}$ is the leading eigenvalue of the loss' Hessian matrix.  
		
		Second, with the assumption of a single output $\hat{y}$, loss presented only at the last time step $T$ and stochastic gradient descent (updates on a single data example as opposed to batch updates), least squares loss in Eq.~\ref{eqn:loss_fcn0} can be simplified to:
		\begin{equation}
			L(W_h) = \frac{1}{2} (\hat y - y)^2,  
		\end{equation}
		resulting in the following difference equations for discrete dynamical systems defined by BPTT (Eq.~\ref{eqn:de_bptt}) and an approximate rule (Eq.~\ref{eqn:de_eprop}):
		\begin{align}
			\left. \Delta W_h \right|_\text{BPTT} &= F(W_h) = -\eta_B (\hat{y}-y) \nabla \hat{y}, \label{eqn:de_bptt2} \\
			\left. \Delta W_h \right|_\text{approximate} &= \hat{F}(W_h) = -\eta_e (\hat{y}-y) \tilde \nabla \hat{y},  \label{eqn:de_eprop2}
		\end{align}
		where we remind the reader that $\tilde \nabla$ is the notation for an approximate gradient. 
		
		We then compute the Jacobian of the difference equations above:
		\begin{align}
			J &= -\eta_B \left( \nabla \hat{y} \nabla \hat{y}^T + (\hat{y}-y) \nabla^2 \hat{y} \right) \quad \text{(for BPTT)}\\ 
			\widehat{J} &= -\eta_e \left( \nabla \hat{y} \tilde \nabla \hat{y} ^T  + (\hat{y}-y) \nabla \tilde \nabla \hat{y} \right) \quad \text{(for an approximate rule)}. \label{eqn:Jacobian_mse}
		\end{align}
		
		In the limit of zero error $(y-\hat{y}) = 0$, i.e. close to an optimum, the term involving $(y-\hat{y})$ becomes negligible. That simplifies the Jacobian to 
		\begin{align}
			J &\approx -\eta_B \nabla \hat{y} \nabla \hat{y} ^T \cr
			\widehat{J} &\approx -\eta_e \nabla \hat{y} \tilde \nabla \hat{y} ^T. \label{eqn:Jacobian_rank1}
		\end{align}
		
		In this case, $J$ and $\widehat{J}$ are \textbf{rank-1} matrices. A rank-1 square matrix has only one nonzero eigenvalue, and by inspection, that one eigenvalue is
		\begin{align}
			\text{For $J$}: |\lambda^J_1(W^*_B)| \quad  &= \left. \eta_B \nabla \hat{y}^T \nabla \hat{y} \right|_{W^*_B} \\
			\rightarrow |\lambda^H| &\overset{\eqref{eqn:Jlam_bptt}}{=} |\lambda^J_1| / \eta_B = \nabla \hat{y}^T \nabla \hat{y} \label{eqn:simplifiedH} \\
			\text{For $\widehat{J}$}: \quad |\widehat{\lambda^J_1}(W^*_e)| &= \left. \eta_e  |\tilde \nabla \hat{y}^T \nabla \hat{y}| \right|_{W^*_e} \cr 
			&\overset{(a)}{=} \left. |\eta_e \rho \nabla \hat{y}^T \nabla \hat{y}|  \right|_{W^*_e} \cr
			&\overset{~\eqref{eqn:simplifiedH}}{=} |\rho \eta_e \lambda^H_1(W^*_e)|, \label{eqn:Jlam_ratio}
		\end{align}
		where equality (a) is explained as follows. We first remind the reader that $\rho$ is defined such that $\vec{\hat{g}} = \rho \vec g + \vec e$ (Eq.~\ref{eqn:rho_def}). For the case of a scalar output $\hat{y}$, $\vec{\hat{g}} = \frac{\partial L}{\partial\hat{y}} \tilde \nabla \hat{y}$ and $\vec g = \frac{\partial L}{\partial \hat{y}} \nabla \hat{y}$. So if we divide both sides of $\vec{\hat{g}} = \rho \vec g + \vec e$ by $\frac{\partial L}{\partial \hat{y}}$ we get $\tilde \nabla \hat{y} = \rho \nabla \hat{y} + \vec e / \frac{\partial L}{\partial \hat{y}}$. Since we have $\vec e \perp \vec g$ by definition, then $\vec e^\top \nabla \hat{y}=0$ because $\vec g$ is just a scaled $\nabla \hat{y}$ when the output is a scalar. This leads to $\tilde \nabla \hat{y}^T \nabla \hat{y} = (\rho \nabla \hat{y} + \vec e / \frac{\partial L}{\partial \hat{y}})^\top \nabla \hat{y} = \rho \nabla \hat{y}^T \nabla \hat{y}$.
		
		Since we assume the gradient descent dynamical system converges to an optimum, this corresponds to an asymptotic stable fixed point. Hence, $|\lambda^J_1|<1$ and $|\widehat{\lambda^J_1}|<1$, which implies:
		\begin{align}
			|\lambda^J_1(W^*_B)| <1  &\overset{\eqref{eqn:Jlam_bptt}}{\rightarrow} \eta_B |\lambda^H_1(W^*_B)| <1  \rightarrow |\lambda^H_1(W^*_B)| < \frac{1}{\eta_B} \\
			|\widehat{\lambda^J_1}(W^*_e)|<1 &\overset{\eqref{eqn:Jlam_ratio}}{\rightarrow} |\rho \eta_e \lambda^H_1(W^*_e)| <1 \rightarrow |\lambda^H_1(W^*_e)| < \frac{1}{|\rho| \eta_e}.
		\end{align}
		
	\end{proof}
	
	\subsection{Discussion on tightness of the bound}
	
	Following the derivation, it is clear that the tightness of the bound will depend on how close the magnitude of leading Jacobian eigenvalue is to $1$ upon convergence. That is related to the distribution of minima flatness along the loss landscape, which impacts the probability of a rule converging to a minima with flatness in a certain range. Such distribution is likely problem dependent. If the loss were convex, there would just be one minimum and the question of minima preference would become irrelevant.
	
	\subsection{Discussion on generality of Theorem~\ref{thm:rank1_eval}} ~\label{scn:thm_gen}
	
	The proof above examines a special case where the Jacobian of weight update equations becomes rank-1. We remind the reader that for the case of multivariate loss, higher rank cases or batch updates, we would not have arrived at the rank-1 Jacobian step (Eq.~\ref{eqn:Jacobian_rank1}). For many tasks considered in neuroscience, the rank 1 case can apply, which explains the validity of Theorem~\ref{thm:rank1_eval}. We also note in passing that for the case of stochastic gradient descent, the Hessian Jacobian correspondence (Eq.~\ref{eqn:Jlam_bptt}) would point to loss' Hessian matrix evaluated on a single example, which could reflect the robustness against perturbing that particular example. Moreover, simulation results show our conclusion holds in higher rank cases (Figure~\ref{fig:LamCurves_rholr}). The challenge of generalizing the proof to higher Jacobian rank case is that we are no longer guaranteed that the leading eigenvectors of BPTT Jacobian coincide with the leading eigenvectors of an approximate rule Jacobian. Thus, it becomes much harder to relate $|\widehat{\lambda^J_1}|$ and $|\lambda^J_1|$. Rather than providing further proof, we provide an intuition for why our conclusion ---  where the convergence behavior between rules differs by their step length along the gradient direction --- can hold in higher rank cases under Assumption~\ref{assume:orth}.
	
	\begin{assume} \label{assume:orth}
		Approximation error vector $\vec e$ (but not $\vec g$) lies orthogonal to the subspace spanned by the leading Hessian eigenvectors. Here, leading Hessian eigenvectors refer to the eigenvectors corresponding to the outlier Hessian eigenvalues in light of the well-known observation that there exists only a few large (outlier) eigenvalues and the rest are near zero~\cite{sagun2016eigenvalues,sagun2017empirical,ghorbani2019investigation}).
	\end{assume}
	The ramification of Assumption~\ref{assume:orth} is that $\vec e$ will lie in the subspace spanned by eigenvectors corresponding to tiny eigenvalues, making $H \vec e$ tiny. In the extreme scenario where $\vec e$ lies in the null space of $H$, $H \vec e$ would be $0$. We verify this assumption numerically in Figure~\ref{fig:overlap}. We saw from the proof above that this assumption is automatically satisfied in the rank-1 Jacobian case. We remark that this assumption should not hold for stochastic gradient noise (SGN), as the SGN covariance matrix is well aligned with the Hessian matrix near a minima~\cite{xie2020diffusion}. This could be why $\vec e$, unlike stochastic gradient noise, does not seem to be contributing much to escaping narrow minima. 
	
	We consider the case of small enough weight updates such that the loss surface can be approximated using second-order Taylor expansion. Thus, the loss change after one update becomes:
	\begin{align}
		\Delta L &\approx \Delta W^T \vec g + \frac{1}{2} \Delta W^T H \Delta W \cr
		&= -\eta_B \vec g^T \vec g + \frac{1}{2} \eta_B^2 \vec g^T H \vec g, \quad \text{(for exact rule)} \\
		\widehat{\Delta L} &\approx \widehat{\Delta W}^T \vec g + \frac{1}{2} \widehat{\Delta W}^T H \widehat{\Delta W} \cr
		&= -\eta_e \vec{\hat{g}}^T \vec g + \frac{1}{2} \eta_e^2 \vec{\hat{g}}^T H \vec{\hat{g}}. \quad \text{(for an approximate rule)}
	\end{align}
	
	We next focus on the first- and second-order Taylor terms ($T_1$ and $T_2$) for the exact rule as well as the terms ($\widehat{T_1}$ and $\widehat{T_2}$) for an approximate rule:
	\[
	T_1:= \eta_B \vec g^T \vec g, \widehat{T_1}:= \eta_e \vec{\hat{g}}^T \vec g, T_2:= \frac{1}{2} \eta_B^2 \vec g^T H \vec g, \widehat{T_2}:= \frac{1}{2} \eta_B^2 \vec{\hat{g}}^T H \vec{\hat{g}},
	\]
	and we note that first and second Taylor terms can determine how likely the update will be trapped in a local minimum:
	\begin{align}
		\Delta L & < 0 \text{ (enables descend)} \quad \rightarrow T_1 > T_2 \cr 
		\Delta L & > 0 \text{ (restricts convergence)} \quad \rightarrow T_1 < T_2 \cr 
		\widehat{\Delta L} & < 0 \text{ (enables descend)} \quad \rightarrow \widehat{T_1} > \widehat{T_2} \cr
		\widehat{\Delta L} & > 0 \text{ (restricts convergence)} \quad \rightarrow \widehat{T_1} < \widehat{T_2}. \label{eqn:T1T2_behavior}
	\end{align}
	
	Given their central role in determining convergence, we compare these terms between exact gradient descent learning and an approximate rule. For the first Taylor term (T1), it is easy to see that: 
	\[
	\vec{\hat{g}}^T \vec g = \rho \vec g^T \vec g.
	\]
	
	For the second Taylor term (T2) and if $H$ is symmetric:
	\begin{align}
		\vec{\hat{g}}^T H \vec{\hat{g}} &= \rho^2 \vec g^T H \vec g + 2 \rho \vec g^T {\underbrace{\textstyle H \vec e  }_{\mathclap{\text{ $\approx 0$ } }}} + \rho^2 \vec e^T {\underbrace{\textstyle H \vec e  }_{\mathclap{\text{$\approx 0$, Assumption~\ref{assume:orth}} }}} \cr
		& \approx \rho^2 \vec g^T H \vec g. 
	\end{align}
	
	To match the convergence behavior between exact gradient descent and an approximate rule (Eq.~\ref{eqn:T1T2_behavior}) on a (locally) second-order loss surface, we can make $(T_1, T_2)$ approximately equal to $(\widehat{T_1}, \widehat{T_2})$ by setting $\eta_B = \rho \eta_e$, which predicts our numerical results (Figure~\ref{fig:LamCurves_rholr}). We note that if Assumption~\ref{assume:orth} were not satisfied, then the above might not hold. We note in passing that if we can satisfy Assumption~\ref{assume:orth} without being near an optimum, then we may not need the negligible training error assumption. 
	
	\newpage 
	
	\section{Additional Simulations}
	
	In the last paragraph in Discussion, we discussed how learning rate modulation could be one of the potential remedies used by the brain. We also explained how learning rate modulation could serve as a balance between the potential benefits of a large learning rate and the numerical stability issue mentioned shortly after the presentation of Figure~\ref{fig:LamCurves_rholr} and Theorem~\ref{thm:rank1_eval} in Results. In Appendix Figure~\ref{fig:LamCurves_lrdecay}, we used a large learning rate early in training to prevent premature stabilization in sharp minima followed by gradual decay to mitigate the stability issue. With this remedy, we observed a reduction in the curvature of the converged solution and an improvement in generalization performance. This result also connects with the finding that sensory depletion during critical periods in training deep networks, which can be related to a small learning rate early in training, can impair learning and yield convergence to sharp minima~\cite{achille2018critical}. However, it is important to note that this strategy does not correct the problem; the gap still exists compared to BPTT, suggesting room for further research. 
	
	\begin{figure}[h!]
		\centering
		\includegraphics[width=0.99\textwidth]{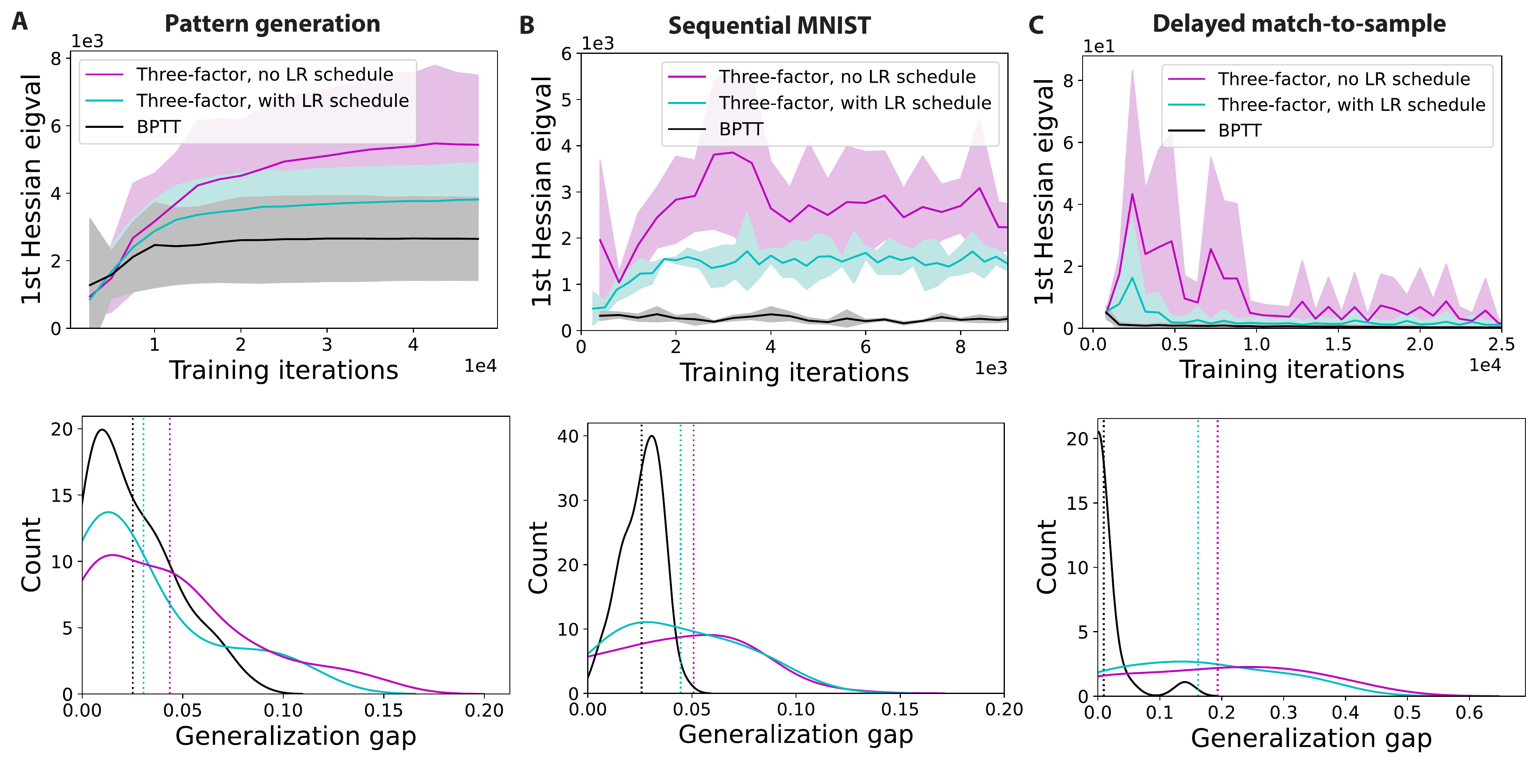}
		\caption{\textbf{Learning rate modulation as a possible remedy of the problem}. We increased the learning rate at the beginning of training to prevent the three-factor rule from stabilizing in sharp minima prematurely, followed by a gradual decay to prevent instability. The top panels show this strategy helps to reduce the curvature of the converged solution. The bottom panels show this leads to a slight improvement in the generalization gap (vertical lines denote distribution mean). However, it is important to note that this strategy does not correct the problem; the gap still exists compared to BPTT, suggesting room for further research. Plotting conventions follow that of the previous figures.
		} 
		\label{fig:LamCurves_lrdecay}
	\end{figure}
	
	\newpage 
	
	We present additional simulations referred to in the main text. In Figure~\ref{fig:LamCurves_rholr}, we attributed the convergence to high curvature regions to reduced along-gradient step length. In Appendix Figure~\ref{fig:gengap_rholr}, we confirm that such high curvature convergence indeed corresponds to worsened generalization performance, thereby linking reduced along-gradient step length to worsened generalization performance. As mentioned in the main text, due to the scale-dependence issue of Hessian spectrum~\cite{dinh2017sharp}, we also used scale-independent measures. For instance, we examined the power-law decay coefficient for the Hessian eigenvalues (Appendix Figure~\ref{fig:EigSpectrum}). We also looked at the recently proposed relative flatness measure~\cite{petzka2021relative} (Appendix Figure~\ref{fig:rel_flat}). These additional measures support the trends observed before: BPTT converges to lower curvature regions compared to the three-factor rule. We also observe that the tendency to approach high curvature regions seems to be a shared problem for temporal truncations of the gradient (Appendix Figure~\ref{fig:LamCurves_trunc}).  
	
	\begin{figure}[h!]
		\centering
		\includegraphics[width=0.99\textwidth]{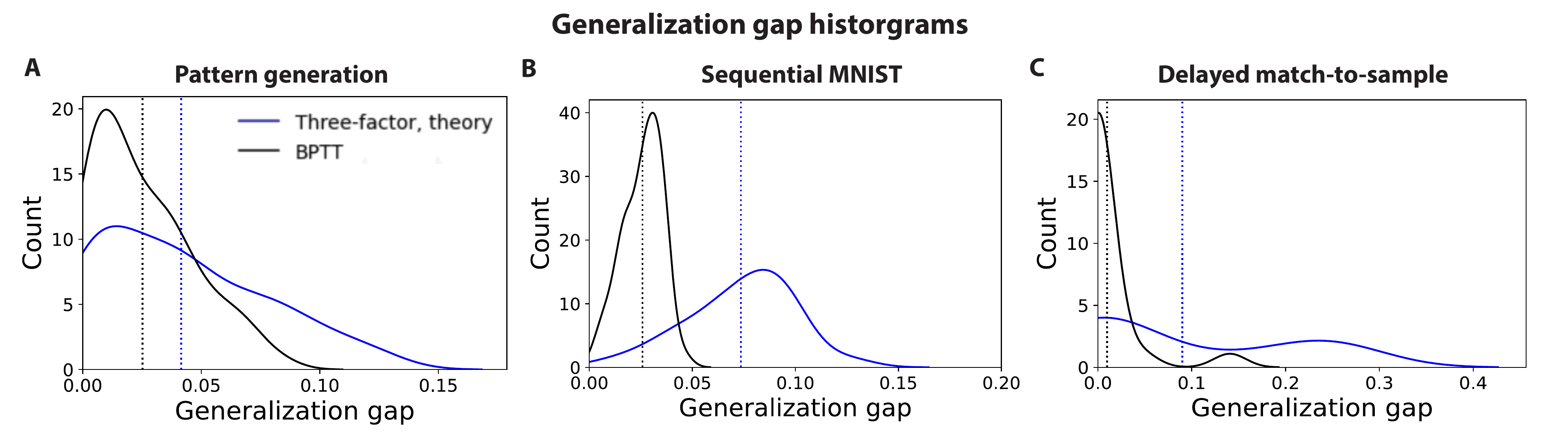}
		\caption{\textbf{Modified BPTT (three-factor, theory) resulted in worse and more variable generalization performance}. Here, we follow the convention of previous generalization gap histogram plots and investigated the generalization performance of modified BPTT (three-factor, theory) in Figure~\ref{fig:LamCurves_rholr}. 
		} 
		\label{fig:gengap_rholr}
	\end{figure}
	
	\begin{figure}[h!]
		\centering
		\includegraphics[width=0.99\textwidth]{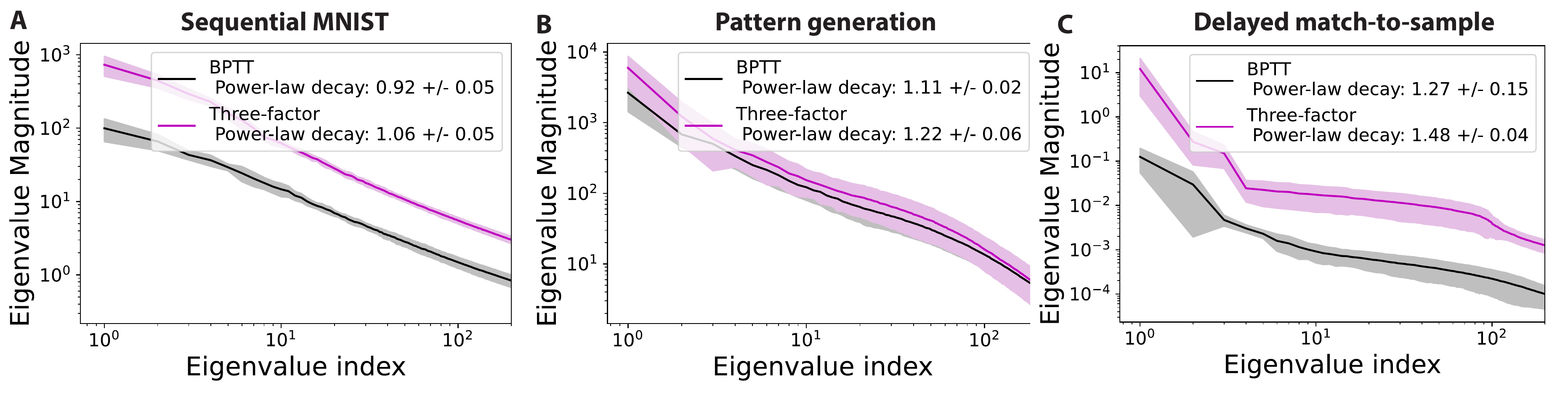}
		\caption{\textbf{Loss' Hessian eigenspectrum for the three-factor rule exhibits significantly steeper power-law decay compared to that of BPTT}. We fit a power-law function to the top 200 eigenvalues at the end of training and measure the decay parameter. Fitting to the top 50 or 100 eigenvalues resulted in similar trends. Solid lines/shaded regions: mean/standard deviation of eigenspectrum obtained at the end of training across five independent runs.
		} 
		\label{fig:EigSpectrum}
	\end{figure}
	
	\begin{figure}[h!]
		\centering
		\includegraphics[width=0.99\textwidth]{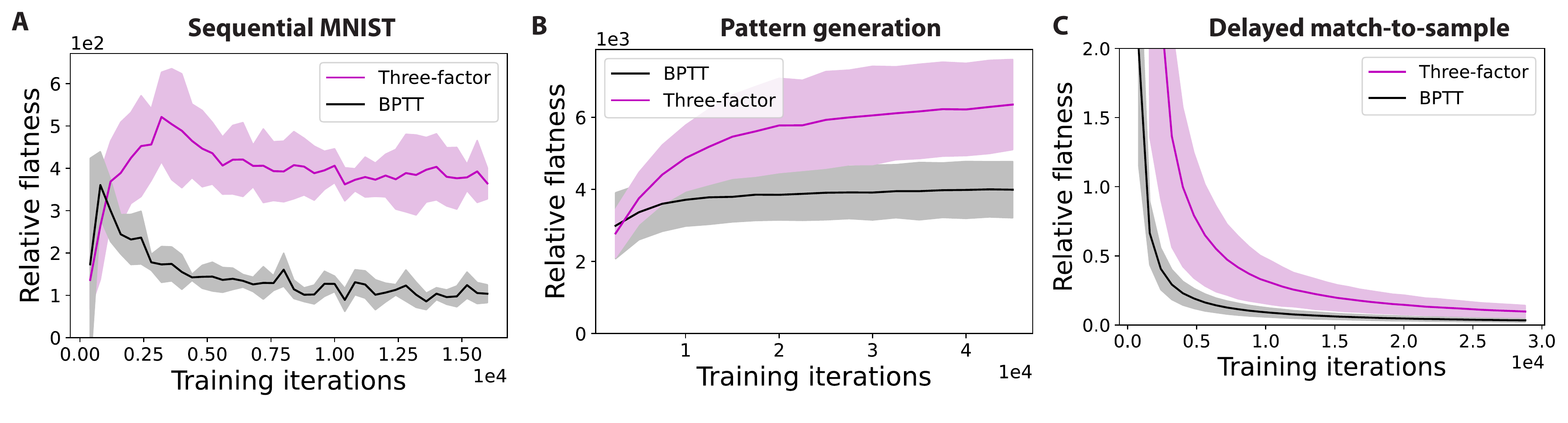}
		\caption{\textbf{Curvature preference behavior corroborated using relative flatness measure~\cite{petzka2021relative}}. Here, the trend is consistent with that of Figure~\ref{fig:LamCurves}. Note that the relative flatness measure can be computationally intensive for recurrent weights, so we computed it for readout weights. Plotting conventions follow that of previous figures. 
		} 
		\label{fig:rel_flat}
	\end{figure}
	
	\begin{figure}[h!]
		\centering
		\includegraphics[width=0.45\textwidth]{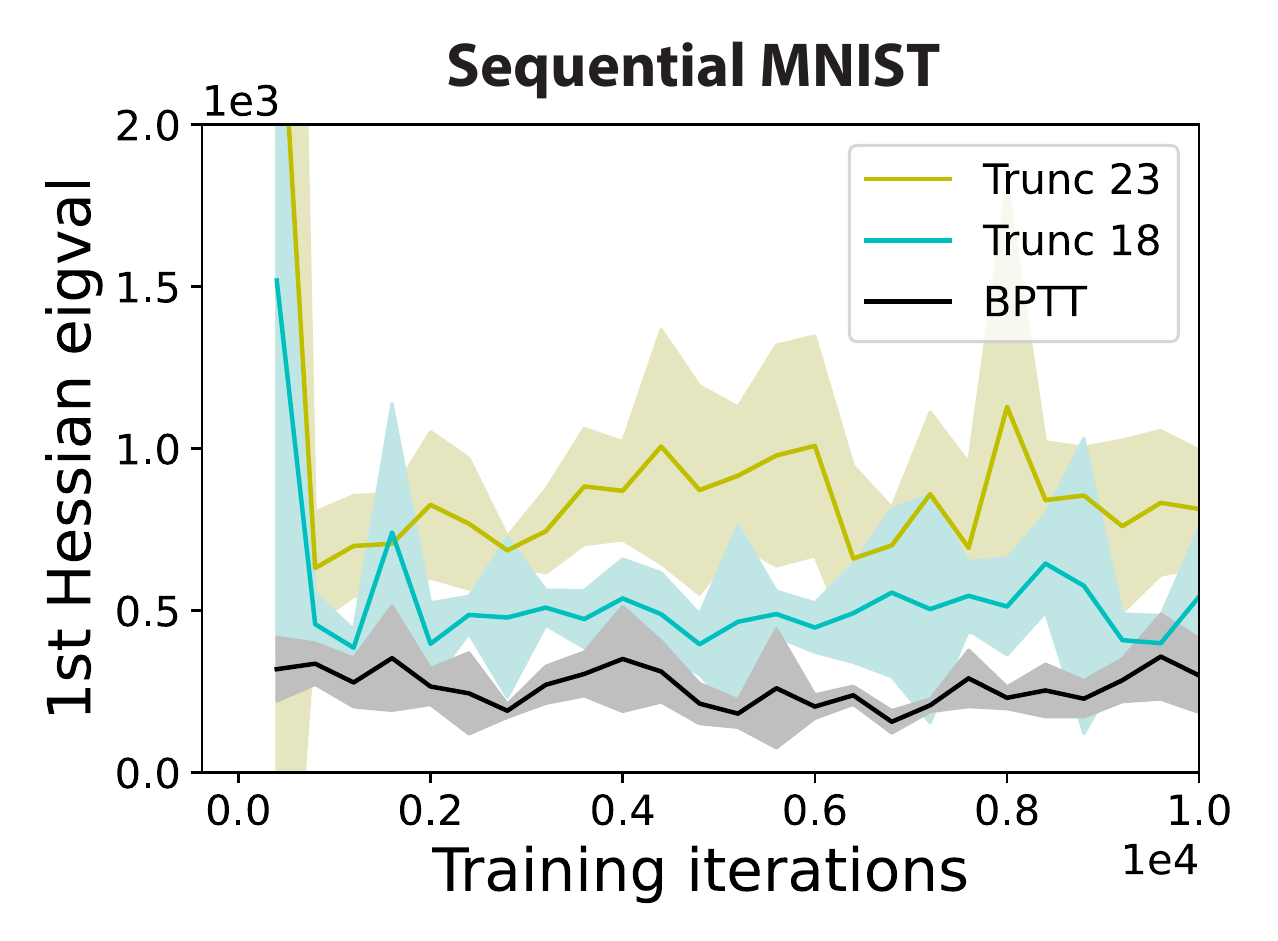}
		\caption{\textbf{Approaching high curvature regions seems to be a shared problem for temporal truncations of the gradient}. We repeat the analysis in Figure~\ref{fig:LamCurves} for truncated BPTT (TBPTT). Here, "Trunc X" means X time steps are truncated during the gradient calculation. We observe that TBPTT tends to converge to high curvature regions. Plotting conventions follow that of previous figures. 
		} 
		\label{fig:LamCurves_trunc}
	\end{figure}
	
\begin{figure}[h!]
		\centering
		\includegraphics[width=0.45\textwidth]{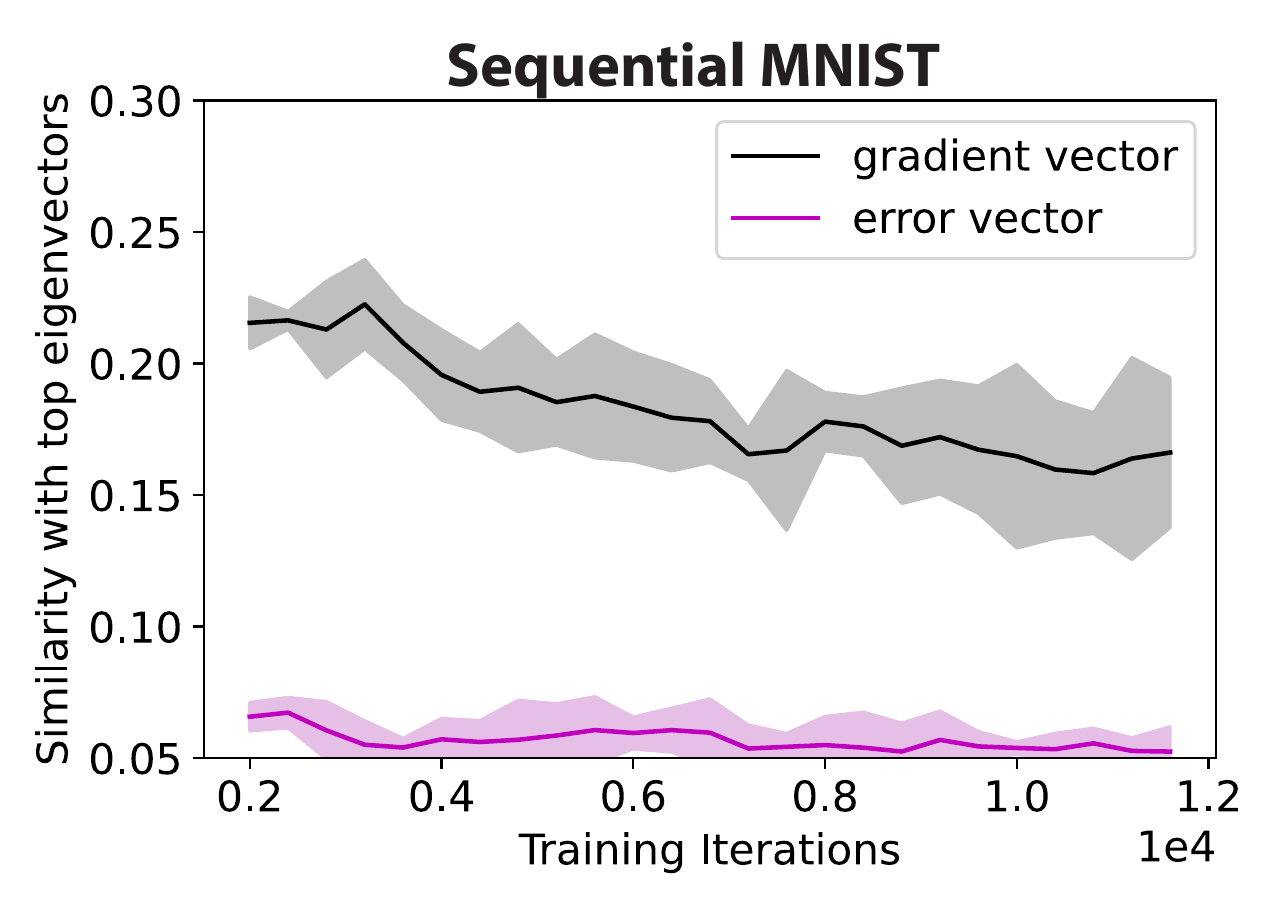}
		\caption{\textbf{Truncation error vector (compared to the gradient) is significantly less aligned with the top Hessian eigenvector subspace}. Following~\cite{jastrzkebski2018relation}, we compute the cosine similarity between the error vector (of the three-factor rule) and a top Hessian eigenvector (averaged over the top 5 eigenvectors). The absolute value of the cosine similarity was taken. We observe weak alignment of the approximation error vector $\vec e$ with the leading Hessian eigenvectors. Similar trends were attained had we averaged over the leading 10 or 20 eigenvectors. Plotting conventions follow that of previous figures. 
		} 
		\label{fig:overlap}
	\end{figure}

	In response to our discussion on the potential impact of noise direction (see explanation shortly after the presentation of Theorem~\ref{thm:rank1_eval}, Discussion section and Appendix~\ref{scn:thm_gen}), we confirm that the error vector $\vec e$ is significantly less aligned with the leading Hessian eigenvectors relative to the gradient vector $\vec g$ (Appendix Figure~\ref{fig:overlap}). As explained in Methods, we used SGD optimizer due to confounding factors in Adam optimizer that could convolute our matching step length analysis in Figure~\ref{fig:LamCurves_rholr}. We observe similar curvature convergence trends as in Figure~\ref{fig:LamCurves} when we repeated the experiments with Adam optimizer in Appendix Figure~\ref{fig:LamCurves_adam}.
	
	\begin{figure}[h!]
		\centering
		\includegraphics[width=0.45\textwidth]{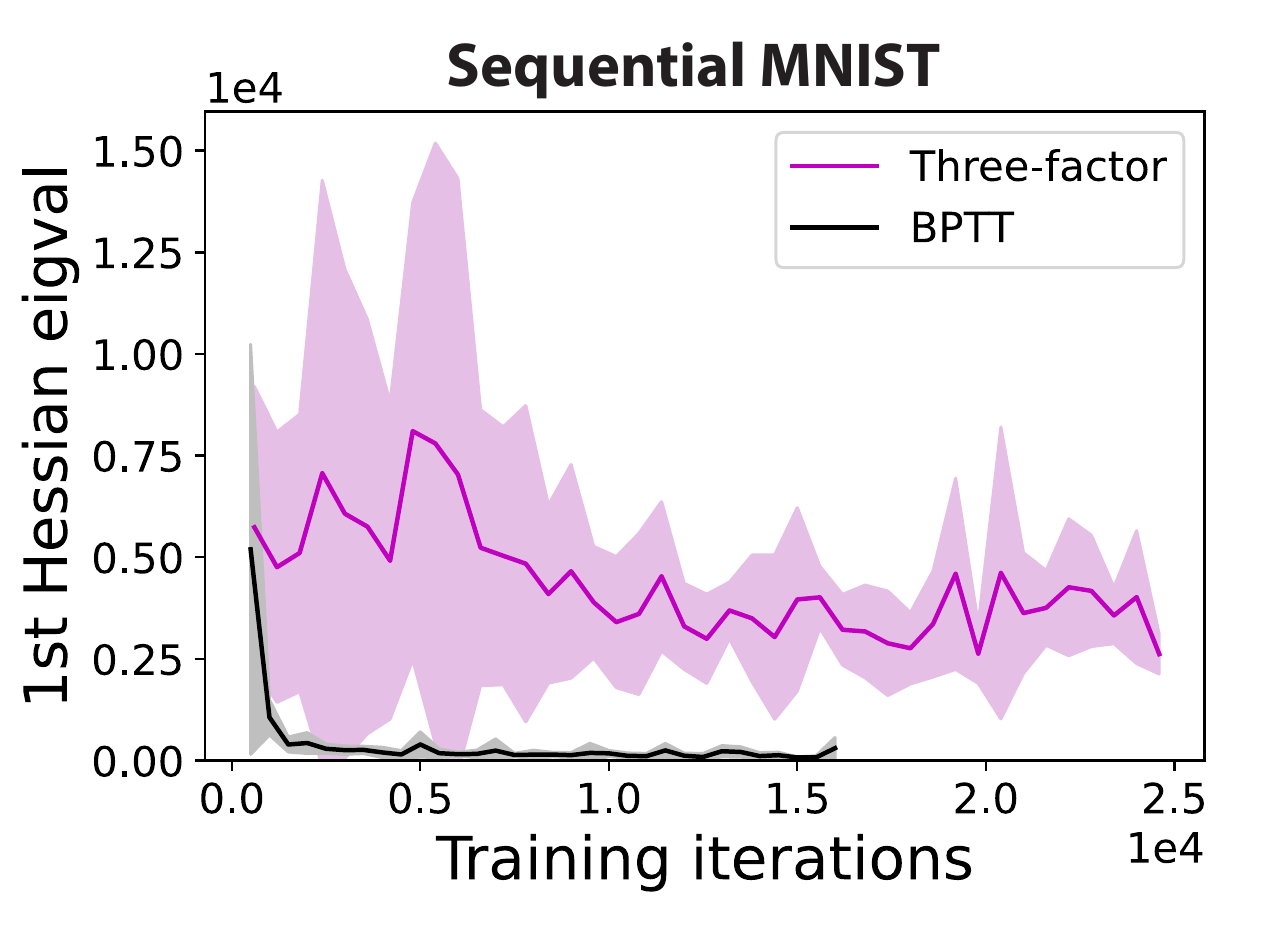}
		\caption{\textbf{Curvature convergence behavior also holds for Adam optimizer}. As explained in Methods, we used SGD optimizer due to confounding factors in Adam optimizer that could convolute our matching step length analysis in Figure~\ref{fig:LamCurves_rholr}. We observe similar trends as in Figure~\ref{fig:LamCurves}. Plotting conventions follow that of previous figures. 
		} 
		\label{fig:LamCurves_adam}
	\end{figure}
	
	\newpage
	
\begin{figure}[h!]
		\centering
		\includegraphics[width=\textwidth]{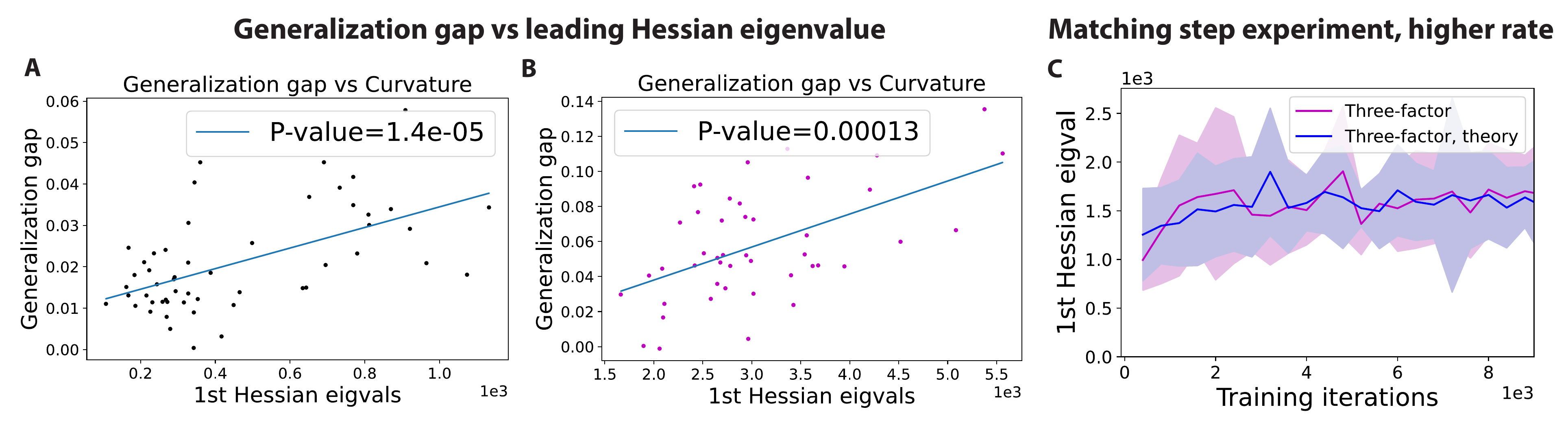}
		\caption{ We repeat the generalization gap vs leading Hessian eigenvalue scatter plot in Figure~\ref{fig:Figure2} with the learning rule fixed for A) BPTT and B) the three-factor rule. As expected, a significant correlation between the generalization gap and leading Hessian eigenvalue is observed. Unlike Figure 2, where the hyperparameters were fixed for each rule (tuned using the procedure in Appendix~\ref{scn:sim_details}), the learning rate is varied here in order to get a wide enough curvature range to observe the correlation. C) The matching step experiments in Figure~\ref{fig:LamCurves_rholr} were repeated here with the learning rate increased by three times for all rules, and the observation agrees with that in Figure~\ref{fig:LamCurves_rholr}. Plotting convention follows that of previous figures. 
		} 
		\label{fig:scatter_and_match}
	\end{figure}

	Finally, to further examine the correlation between leading Hessian eigenvalue and generalization performance (observed in Figure~\ref{fig:Figure2}), we also observed such correlation correlation for runs with the learning rule fixed (Appendix Figure~\ref{fig:scatter_and_match}). For the matching step experiment in Figure~\ref{fig:LamCurves_rholr}, similar observations were also made when we repeated the experiment at three times the learning rate (Appendix Figure~\ref{fig:scatter_and_match}C). Moreover, we stopped BPTT early to match the test accuracy of the three-factor rule, and observed similar curvature convergence and generalization performance trends as previously (Appendix Table~\ref{table:1}).
	
	\begin{table}[h!]
		\centering
		\begin{tabular}{||c c c c||} 
			\hline
			Learning & Leading Hessian eigenvalue & Generalization gap \\ [0.5ex] 
			\hline\hline
			Three-factor & $2550 \pm 490$ & $0.5 \pm 0.3$ \\ 
			BPTT, early stopping & $316 \pm 84$ & $0.2 \pm 0.1$ \\ [1ex] 
			\hline
		\end{tabular}
		\caption{BPTT stopped early to match the test accuracy of the three-factor rule for the sequential MNIST task. Higher generalization gap and leading Hessian eigenvalue is again observed for the three-factor rule, as expected. Each rule is repeated for five runs with different random weight initialization. }
		\label{table:1}
	\end{table}
	
\end{document}